\documentclass{article}

\usepackage[preprint]{neurips_2026}

\usepackage[utf8]{inputenc} 
\usepackage[T1]{fontenc}    
\usepackage{hyperref}       
\usepackage{url}            
\usepackage{booktabs}       
\usepackage{amsfonts}       
\usepackage{multirow}
\usepackage{amsmath}
\usepackage{nicefrac}       
\usepackage{microtype}      
\usepackage{xcolor}         
\usepackage{graphicx}
\usepackage{enumitem}
\usepackage{booktabs}
\usepackage{makecell}
\usepackage{colortbl}
\usepackage{xcolor}
\usepackage{subcaption}
\usepackage{wrapfig}
\usepackage{caption}
\usepackage{subcaption}
\usepackage{booktabs}
\usepackage{graphicx}
\usepackage{hyperref}
\usepackage{fontawesome5}
\usepackage{xcolor}

\definecolor{hfyellow}{HTML}{FFD21E}

\newcommand{\hficon}{%
    \raisebox{-0.15ex}{\colorbox{hfyellow}{\textbf{\scriptsize HF}}}%
}

   \usepackage[T1]{fontenc}
   \usepackage{amsmath,amssymb,amsthm}
   \usepackage{xcolor}
   \usepackage[most]{tcolorbox}     
   \usepackage{listings}
   \usepackage{enumitem}
   \usepackage{microtype}
   \usepackage{pifont}              
   \usepackage{hyperref}            
\definecolor{todo}{RGB}{196, 30, 58}
\newcommand{\name}{MetaAgent-X\,}
\makeatletter
\def\blfootnote{\xdef\@thefnmark{}\@footnotetext}
\makeatother
\title{\name: Breaking the Ceiling of Automatic Multi-Agent Systems via End-to-End Reinforcement Learning}

\author{%
  \vspace{-25pt}\\
  \textbf{Yaolun Zhang$^{1,5,*}$,\quad Yujie Zhao$^{2,*}$ } \\
  \textbf{Nan Wang$^{3,\dagger}$, \quad Yiran Wu$^{4,5}$,\quad Jiayu Chang$^{2}$, \quad Yizhao Chen$^{2}$ } \\
  \textbf{Qingyun Wu$^{4,5}$, \quad Jishen Zhao$^{2}$, \quad Huazheng Wang$^{1,5}$}\vspace{3pt} \\
  $^{1}$Oregon State University \quad\quad $^{2}$UCSD \\
  $^{3}$Amazon AGI \quad\quad $^{4}$Pennsylvania State University \quad\quad $^{5}$AG2AI, Inc.\vspace{3pt} \\
  \texttt{\small \{zhanyaol, huazheng.wang\}@oregonstate.edu} \\
\texttt{\small \{yuz285,yic138,jzhao\}@ucsd.edu} \\
\texttt{\small nanww@amazon.com}
  \texttt{\small \quad \{yiran.wu, qingyun.wu\}@psu.edu}\vspace{3pt}
}

\begin{document}
\maketitle 
\blfootnote{$^{*}$Equal contribution.}
\blfootnote{$^{\dagger}$This work is unrelated to the author's position at Amazon.}

\begin{center}
\vspace{-1em}
\href{https://github.com/pettingllms-ai/PettingLLMs}{\faGithub\ \texttt{Code}} \quad\textbar\quad
\href{https://huggingface.co/Mercury7353/MetaAgent-X}{\hficon\ \texttt{Model}}
\end{center}
\begin{abstract}
Automatic multi-agent systems aim to instantiate agent workflows without relying on manually designed or fixed orchestration. However, existing automatic MAS approaches remain only partially adaptive: they either perform training-free test-time search or optimize the meta-level designer while keeping downstream execution agents frozen, which creating a frozen-executor ceiling and leaving the end-to-end training of self-designing and self-executing agentic models unexplored. To address this, we introduce \textbf{\name}, an end-to-end reinforcement learning framework that jointly optimizes automatic MAS design and execution. \name enables script-based MAS generation, execution rollout collection, and credit assignment for both designer and executor trajectories. To support stable and scalable optimization, we propose Executor Designer Hierarchical Rollout and Stagewise Co-evolution to improve training stability and expose the dynamics of designer-executor co-evolution. \name consistently outperforms existing automatic MAS baselines, achieving up to 21.7\% gains. Comprehensive ablations show that both designer and executor improve throughout training, and that effective automatic MAS learning follows a stagewise co-evolution process. These results establish end-to-end trainable automatic MAS as a practical paradigm for building self-designing and self-executing agentic models.
\end{abstract}

\section{Introduction}


Multi-agent systems (MAS) have demonstrated clear advantages over single-agent approaches across a wide range of domains, including medical decision-making~\citep{kim2024mdagents,zhou2025mam}, scientific discovery~\citep{su2024many,ghafarollahi2024sciagents}, financial trading~\citep{xiao2024tradingagents}, software engineering~\citep{yu2025orcaloca,hong2023metagpt,chen2024coder}, and hardware design~\citep{zhao2024mage,ho2025marco}. Rather than relying on manually specified or fixed workflows, recent work has increasingly turned to meta-agents as a paradigm for automatically designing and instantiating the multi-agent system flow best suited to each task, enabling more adaptive orchestration and execution of MAS~\citep{gao2025flowreasoner,ye2025masgpt,dang2025evolving,nielsen2025conductor,zhang2025metaagentautomaticallyconstructingmultiagent}. 

Meanwhile, as agentic reinforcement learning and self-evolving paradigms have emerged as promising pathways to transform large language models into interactive, continuously improving decision-makers~\citep{wang2025ragen, cheng2025agentr1, li2025intheflowagenticoptimizationeffective, zhao2026strongermasmultiagentreinforcementlearning, zhang2026evaefficientreinforcementlearning, xia2025agent0, chen2025dreamgym, fu2025evolver}, recent automatic MAS begin 
to embrace these paradigms, their transition remains incomplete. Current approaches typically restrict adaptation to non-training test-time search, or only optimize the MAS designer, while freezing downstream execution agents \citep{ye2025masgpt, gao2025flowreasoner, dang2025evolving, nielsen2025conductor, wang2025mas2selfgenerativeselfconfiguringselfrectifying}. 
Yet, 
end-to-end training of self-designing and 
self-executing auto-MAS remains unexplored, 
resulting in two 
fundamental limitations:
1) \textbf{Parameter-level disjunction.} Existing methods couple the designer and executor only through prompt-level interactions at inference time, without optimization signals that update the underlying policy based on downstream execution outcomes. As a result, a frozen executor imposes a hard ceiling on the meta-designer, while the designer cannot induce specialized execution behaviors from its counterpart.
2) \textbf{Vague co-evolution dynamics.} The dynamics by which 
designer and executor could co-evolve under joint training, and where 
each role's improvement remains unclear in practice and in understanding the mechanism.

\begin{figure}
    \centering
    \includegraphics[width=0.8\linewidth]{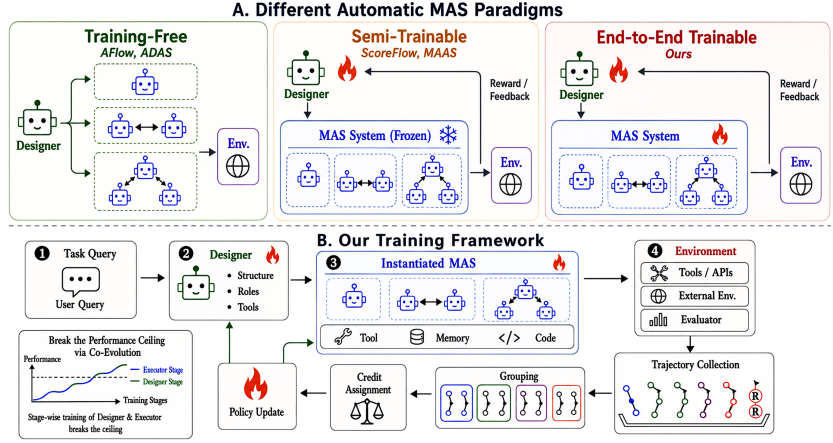}
    \caption{
\textbf{From Partial Adaptation to End-to-End Trainable Automatic MAS.}
\textbf{A.} Comparison of three automatic MAS paradigms. 
\textbf{B.} Overview of our training framework.
}
\label{fig:automatic_mas_paradigms}
    \label{fig:meta_agent_paradigms}
    \vspace{-20pt}
\end{figure}

As shown in Figure~\ref{fig:automatic_mas_paradigms} (A), existing automatic MAS approaches remain partially adaptive: they either search over MAS structures at test time or optimize only the designer while freezing the execution system. To overcome these limitations, we introduce \textbf{\name}, an end-to-end framework 
to train agentic models that can self-design and self-execute MAS.
Figure~\ref{fig:automatic_mas_paradigms}(B) gives an overview of \name, where task-conditioned auto MAS designs are instantiated, executed, grouped, and collected for role-aware policy updates.
To address the first limitation, \name facilitates script-based MAS generation, rollout collection, and precise credit assignment for both the designer and the executor. To address the second limitation, the framework incorporates diverse evolving mechanisms, such as hierarchical rollouts and stage-wise optimization, allowing us to isolate the critical decision factors that drive auto-MAS co-evolution.

Our framework  consists of three novel design principles.
First, \name supports flexible designer executor optimization across tasks and domains, where the two components can be trained with diverse evolving mechanisms. This flexibility enables a systematic analysis of how designer-executor co-evolution emerges and how each component contributes to the final automatic MAS capability. Second, we propose \textbf{Executor-Designer Hierarchical Rollout}, which organizes the interaction process as a two-level tree structure to support efficient rollout generation and accurate credit assignment. Third, we propose \textbf{Stagewise Co-evolution}, which decouples the learning stages of the designer and executor to improve training stability and scalability. Based on these mechanisms, we conduct comprehensive experiments and ablation studies to evaluate the effectiveness of \name and analyze the internal dynamics of designer-executor co-evolution. Across six math and code benchmarks and two different base models, \name outperforms the baselines by up to 21.7\%.

This paper makes the following contributions:
\begin{enumerate}
    \item We propose \textbf{\name}, an end-to-end training framework for automatic MAS, which explicitly optimizes designer and executor agents together.

    \item We introduce two mechanisms for stable and scalable meta agent optimization: (i) \textbf{Executor Designer Hierarchical Rollout}, which enables structured rollout generation and accurate credit assignment and (ii) \textbf{Stagewise Co-evolution}, which supports decoupled and scalable designer executor learning.

    \item We demonstrate that \name achieves consistent gains across diverse math and code benchmarks, surpassing both single agent and automatic MAS baselines by up to 21.7\%

    \item We conduct comprehensive ablation studies to examine the internal mechanisms of meta-agent co-evolution. Our analysis shows that (1) both the designer and the executor are optimized throughout training across tasks and domains, and (2) such effective co-evolution follows a stagewise process in which the two components benefit from decoupled optimizations.
\end{enumerate}

\section{Related work}

\subsection{Meta Agents for Automatic Multi-Agent Systems}
LLM-based MAS improve complex problem solving by decomposing tasks into specialized roles, structured interactions, and coordination protocols \citep{qian2023chatdev,hong2024metagpt,wu2023autogen}. Beyond manually designed workflows, recent work introduces meta-agent that automatically constructs or adapts an executable MAS for each input task \citep{ye2025masgpt,gao2025flowreasoner,dang2025evolving,nielsen2025conductor,zhang2025metaagentautomaticallyconstructingmultiagent}. A meta-agent maps a query into roles, prompts, communication patterns, or execution flows, after which the instantiated system interacts with the environment to produce the final outcome.

As shown in Fig.~\ref{fig:meta_agent_paradigms}, existing automatic MAS methods mainly fall into two partial adaptation regimes. Training-free adaptation searches over prompts, roles, workflows, or agent organizations at test time without updating model parameters \citep{zhang2025metaagentautomaticallyconstructingmultiagent,dang2025evolving}. Semi-trainable adaptation optimizes a meta-level designer or controller while keeping downstream executors fixed. Examples include MAS-GPT\citep{ye2025masgpt}, which generates query-adaptive MAS designs, FlowReasoner\citep{gao2025flowreasoner}, which learns query-level multi-agent reasoning flows, and orchestration-based controllers for dynamic coordination \citep{nielsen2025conductor}. Also, MAS$^2$~\citep{wang2025mas2selfgenerativeselfconfiguringselfrectifying} trains the designer via reinforcement learning while keep using api-based models as executors. These methods improve system design or orchestration, but do not jointly optimize executor policies.

This partial adaptation limits automatic MAS because frozen executors impose a ceiling on final performance and prevent designer executor co-adaptation. Chain-of-Agents takes a related end-to-end direction by training an Agent Foundation Model through multi-agent distillation and agentic reinforcement learning \citep{li2025chainofagents}, but largely optimizes the agent system as a unified behavior and treat MAS as a simple chain of thought without context management. In contrast, our work studies the end-to-end trainable regime, where automatic MAS evolves both how agent systems are designed and how instantiated agents execute them, making designer executor co-evolution explicit and analyzable.
\subsection{Agent System Self Evolution and Multi-Agent Training}

In parallel with meta-agent based automatic MAS, agentic reinforcement learning and self evolution have emerged as promising paradigms for improving LLM agents through interaction, environment feedback, and iterative experience collection \citep{wang2025ragen,cheng2025agentr1,li2025intheflowagenticoptimizationeffective,zhao2026strongermasmultiagentreinforcementlearning,zhang2026evaefficientreinforcementlearning,xia2025agent0,chen2025dreamgym,fu2025evolver}. Within the multi-agent setting, recent methods such as MAPoRL \citep{park2025maporl}, AT-GRPO \citep{zhao2026strongermasmultiagentreinforcementlearning}, Dr. MAS \citep{feng2026drmas}, MAE \citep{chen2025multiagentevolvellmselfimprove}, and MARFT \citep{liao2025marftmultiagentreinforcementfinetuning} mainly focus on improving collaboration under fixed or predefined multi-agent structures. These methods study important problems such as multi-agent credit assignment, coordination, communication, and training stability. However, the agent organization itself is usually treated as given, rather than as a learned object that should be generated, evaluated, and improved together with execution behavior.

Our work differs from these self evolution and agent foundation model approaches in both objective and analysis. Instead of assuming a fixed MAS structure or optimizing an agent system as an undifferentiated whole, we explicitly formulate automatic MAS learning as a designer-executor co-evolution problem.  This enables us to break the frozen-executor performance ceiling while also studying the internal mechanism of automatic MAS co-evolution.
\section{Method}
\subsection{End to End Online Meta Agent RL Pipeline}
\label{sec:online_pipeline}

\begin{figure}
    \centering
    \includegraphics[width=1\linewidth]{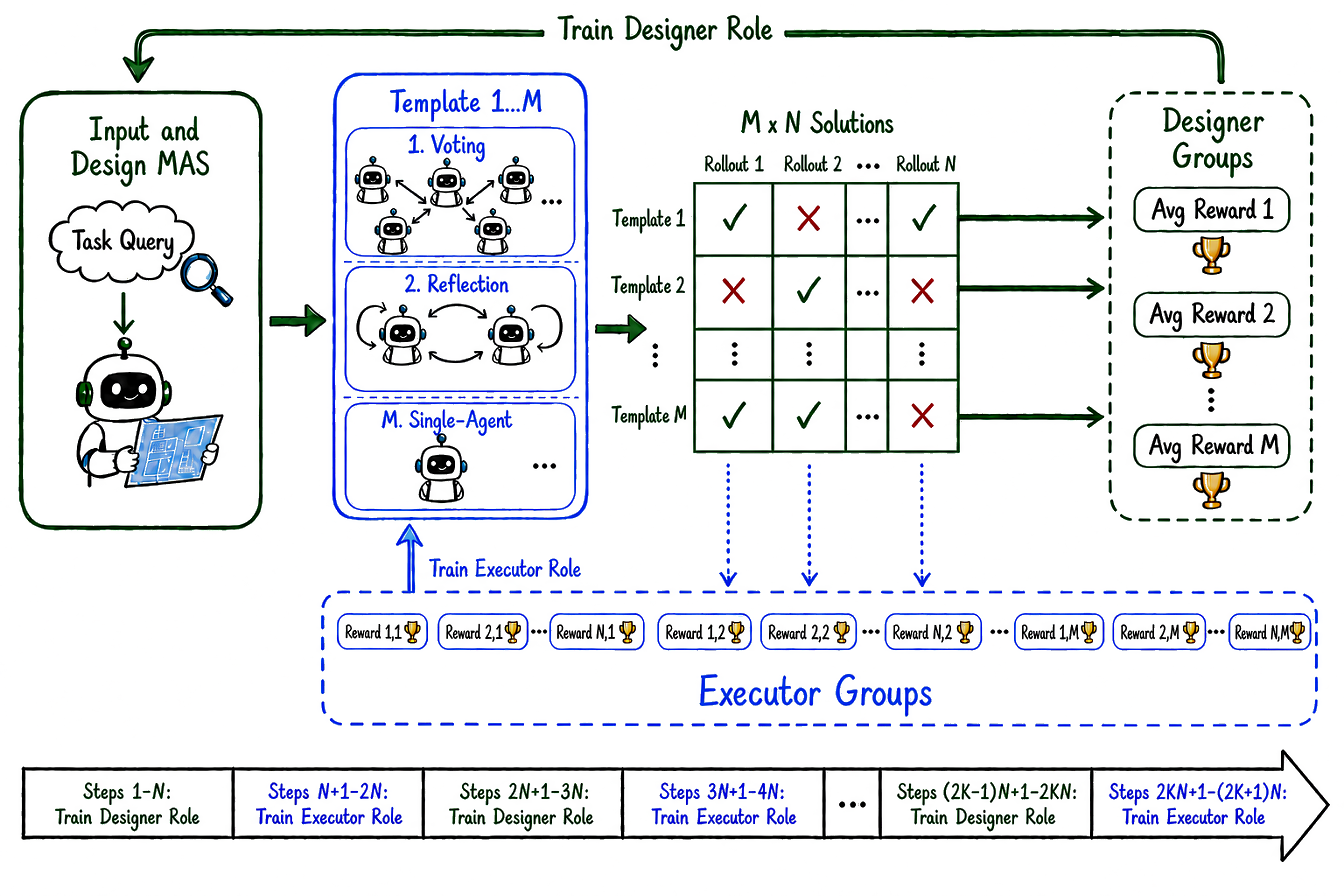}
    \caption{Overview of the end to end online \name pipeline. The  Designer first generate a task-specific multi agent system, then the Executor run the instantiated MAS in the environment. The collected trajectories and rewards are labeled by role and optimized with GRPO.}
    \label{fig:online_pipeline}
\end{figure}

Figure~\ref{fig:online_pipeline} shows our reinforcement learning pipeline. Given a task query $q$, the MetaAgent first uses a Designer policy
$\pi_{\vartheta_\mathcal{D}}^{\mathcal{D}}$ to generate a task specific
multi agent system, and then uses an Executor policy
$\pi_{\vartheta_\mathcal{E}}^{\mathcal{E}}$ to run the instantiated system
in an external environment. We denote the full trainable parameter set by
$\vartheta = \{\vartheta_\mathcal{D}, \vartheta_\mathcal{E}\}$. This notation
covers both policy sharing and policy splitting. In the shared policy setting,
$\vartheta_\mathcal{D} = \vartheta_\mathcal{E} = \theta$; in the split policy
setting, $\vartheta_\mathcal{D}$ and $\vartheta_\mathcal{E}$ are optimized as
separate parameter sets. The learning problem is therefore a coupled online reinforcement learning
problem:
\begin{equation}
    d \sim \pi_{\vartheta_\mathcal{D}}^{\mathcal{D}}(\cdot \mid q),
    \qquad
    e \sim \pi_{\vartheta_\mathcal{E}}^{\mathcal{E}}(\cdot \mid q, d),
    \qquad
    R = R(q, d, e),
    \label{eq:online_problem}
\end{equation}
where $d$ denotes the generated system design, $e$ denotes the execution
trajectory, and $R$ is the environment feedback returned after execution.
The central challenge is that design and execution are interdependent; their
performance is coupled. Thus, the training pipeline must support online system
construction, batched environment execution, trajectory collection, and
role aware credit assignment within a unified RL framework.

\paragraph{Online system construction.}
To support compositional system design, we build a training framework contains predefined coordination structures, agent templates, and tool interfaces.
For each query, the Designer composes these building blocks into a customized multi agent system by generating lightweight Python scripts.
These scripts specify the agent roles, interaction protocol, tool usage pattern, and execution control flow. After a design is instantiated, the Executor runs the generated workflow in the target environment. Our framework supports batched rollout execution across multiple queries and sampled designs. For each rollout, the system records the rollouts, environment observations, tool calls, and the outcome-based rewards (detailed in Appendix~\ref{sec:reward-design}).

\paragraph{GRPO objective.}
We optimize the role policies with Group Relative Policy Optimization(GRPO).
For each role $r \in \{\mathcal{D}, \mathcal{E}\}$, let
$\mathcal{G}^{r}$ denote the corresponding GRPO group, and let
$\hat{A}^{r}_i$ be the normalized role specific advantage for trajectory $i$.
Let $\vartheta_r$ denote the parameters used by role $r$. The clipped policy
objective for role $r$ is
\begin{equation}
    \mathcal{L}_{r}(\vartheta_r)
    =
    -
    \frac{1}{|\mathcal{G}^{r}|}
    \sum_{i \in \mathcal{G}^{r}}
    \min
    \left(
        \rho_i^{r}(\vartheta_r) \hat{A}^{r}_i,
        \operatorname{clip}
        \left(
            \rho_i^{r}(\vartheta_r),
            1 - \epsilon,
            1 + \epsilon
        \right)
        \hat{A}^{r}_i
    \right),
    \label{eq:grpo_role_loss}
\end{equation}
where
\begin{equation}
    \rho_i^{r}(\vartheta_r)
    =
    \frac{
        \pi_{\vartheta_r}^{r}(o_i \mid c_i)
    }{
        \pi_{\vartheta_{r,\mathrm{old}}}^{r}(o_i \mid c_i)
    }.
    \label{eq:grpo_ratio}
\end{equation}
Here $c_i$ is the context of trajectory $i$, $o_i$ is the generated output
tokens, and $\pi_{\vartheta_{r,\mathrm{old}}}^{r}$ is the role specific
behavior policy used for rollout collection. The role specific advantages
$\hat{A}^{\mathcal{D}}$ and $\hat{A}^{\mathcal{E}}$ are computed using the
hierarchical credit assignment scheme in Section~\ref{sec:tree}.

Further, because the Designer and Executor are optimized through coupled online feedback, we introduce a stagewise training schedule that provides a relatively stable environment for optimizing both roles. We discuss the details in Section~\ref{sec:stagewise}.
\subsection{Hierarchical Credit Assignment via Tree-Structured Rollout}
\label{sec:tree}

A central challenge in training end-to-end automatic MAS with RL is \textbf{credit assignment}: when a multi-agent system succeeds or fails at a task, is the outcome attributable to the quality of the Designer's plan or the competence of the Executor's actions?
Standard single-level rollout conflates these two sources of variation, producing entangled reward signals that destabilize training.
We address this through a tree-structured rollout scheme that decomposes credit across roles.

\paragraph{Bi-level Tree-Structured Rollout.}
For each training question $q$, we construct a two-level sampling tree.
At the first level, the Designer $\pi_{\vartheta_\mathcal{D}}^{\mathcal{D}}$ generates $M$ independent multi-agent system designs $\{d_1, d_2, \ldots, d_M\}$, each specifying a distinct agent topology, role assignment, and coordination protocol.
At the second level, for each design $d_i$, the Executor $\pi_{\vartheta_\mathcal{E}}^{\mathcal{E}}$ carries out $N$ independent execution rollouts $\{e_{i,1}, e_{i,2}, \ldots, e_{i,N}\}$.
This yields an $M \times N$ evaluation matrix per question, where entry $(i, j)$ corresponds to design $d_i$ executed by rollout $e_{i,j}$, with outcome reward $R(e_{i,j}, d_i)$.

\paragraph{Decomposed Advantage Estimation.}
The tree structure enables us to compute \emph{separate} advantage estimates for each role via distinct grouping strategies within the GRPO framework.

\textit{Designer advantage.}
To isolate the effect of design quality from execution-level stochasticity, we aggregate over the execution level.
For each design $d_i$ under question $q$, we define the design-level reward as the mean execution outcome:
\begin{equation}
    \bar{R}^{\mathcal{D}}_i = \frac{1}{N} \sum_{j=1}^{N} R(e_{i,j}, d_i).
    \label{eq:reward_d}
\end{equation}
The advantage for design $d_i$ is then computed by comparing against all $M$ designs for the same question:
\begin{equation}
    \hat{A}^{\mathcal{D}}_i = \frac{\bar{R}^{\mathcal{D}}_i - \mu_q^{\mathcal{D}}}{\sigma_q^{\mathcal{D}} + \epsilon},
    \quad \text{where} \quad
    \mu_q^{\mathcal{D}} = \frac{1}{M}\sum_{k=1}^{M} \bar{R}^{\mathcal{D}}_k,
    \quad
    \sigma_q^{\mathcal{D}} = \text{std}(\{\bar{R}^{\mathcal{D}}_k\}_{k=1}^{M}).
    \label{eq:adv_d}
\end{equation}
By averaging over $N$ executions, the stochasticity of individual rollouts is smoothed out, yielding a reward signal that reflects the intrinsic quality of the design itself.

\textit{Executor advantage.}
For each execution rollout $e_{i,j}$, the Executor produces a set of agent trajectories, denoted by  $\mathcal{T}_{i,j}$.
We use the outcome reward of the rollout, $R(e_{i,j}, d_i)$, as the reward for all trajectories in $\mathcal{T}_{i,j}$.
To compute the Executor advantage, we collect all executor trajectories for the same question into a GRPO group:
\begin{equation}
    \mathcal{G}_q^{\mathcal{E}} =
    \left\{ \tau \;\middle|\; \tau \in \mathcal{T}_{i,j},\; i \in [M],\; j \in [N] \right\}.
    \label{eq:group_e}
\end{equation}
The advantage of each trajectory is then normalized at the question level:
\begin{equation}
    \hat{A}^{\mathcal{E}}(\tau) =
    \frac{R(e_{i,j}, d_i) - \mu_q^{\mathcal{E}}}{\sigma_q^{\mathcal{E}} + \epsilon},
    \quad \tau \in \mathcal{T}_{i,j},
    \label{eq:adv_e}
\end{equation}
where $\mu_q^{\mathcal{E}}$ and $\sigma_q^{\mathcal{E}}$ denote the mean and standard deviation of the rollout rewards associated with trajectories in $\mathcal{G}_q^{\mathcal{E}}$.
Compared with single-level rollout normalization, question-level normalization compares executor trajectories generated under both the same and different designs, thereby providing a more stable training signal for the executor.

\subsection{Stagewise Executor-Designer Co-evolution}
\label{sec:stagewise}

The hierarchical rollout in Section~\ref{sec:tree} provides decomposed reward signals for the Designer ($\mathcal{D}$) and Executor ($\mathcal{E}$) roles.
However, since the two roles' rewards are mutually conditioned, a fundamental optimization challenge arises: \emph{how should we update $\pi_\theta$ when $\mathcal{D}$ and $\mathcal{E}$ serve as each other's environment?}

The Designer and Executor form a tightly coupled system where each role is the other's environment: the Executor acts within the MAS structure emitted by the Designer, while the Designer's reward is decided by the capability of the Executor.
Formally, the return is a nested expectation:
\begin{equation}
    J(\theta) = \mathbb{E}_{d \sim \pi_{\vartheta_\mathcal{D}}^{\mathcal{D}}} \left[ \mathbb{E}_{e \sim \pi_{\vartheta_\mathcal{E}}^{\mathcal{E}}(\cdot \mid d)} \left[ R(e, d) \right] \right],
    \label{eq:bilevel}
\end{equation}
 Inspired by
multi-agent RL studies on non-stationarity and sequential optimization
\citep{hernandezleal2019surveylearningmultiagentenvironments,
yu2022surprisingeffectivenessppocooperative,
nekoei2023dealingnonstationaritydecentralizedcooperative}, we introduce a
stagewise schedule that alternates which role provides the trajectories for
policy-gradient updates. At training step $t$, we select the active role by fixed-length phases of
$K$ steps:
\begin{equation}
    (\alpha_\mathcal{D}^{(t)},\; \alpha_\mathcal{E}^{(t)}) =
    \begin{cases}
        (0,\; 1), & \lfloor t / K \rfloor \bmod 2 = 0
        \quad \text{Executor stage}, \\
        (1,\; 0), & \lfloor t / K \rfloor \bmod 2 = 1
        \quad \text{Designer stage}.
    \end{cases}
    \label{eq:stagewise}
\end{equation}
Only trajectories from the active role contribute to the gradient, while the
shared parameters $\vartheta$ are updated continuously. This isolates each
phase to one reward distribution and reduces gradient interference between
role-specific objectives.

The two stages form a co-evolutionary loop. Executor stages improve the
ability to solve tasks under the current design distribution, producing more
reliable execution outcomes. Designers then use these lower-noise
returns to learn structures that better exploit the improved Executor. As a result, the effective reward
distribution becomes non-stationary and the two role-specific objectives can
produce noisy or conflicting updates.

\section{Experiment}
\subsection{Experiments Setup}
\label{sec:exp-setting}

\paragraph{Models and Compute.}
We train and evaluate Qwen3~\citep{Qwen3TR} at the 4B and 8B parameter scales in no-thinking mode. All experiments are conducted on a single node equipped with eight H200 GPUs. Unless otherwise specified, both the maximum prompt length and maximum response length are set to $8192$ tokens. We use the shared-policy setting, in which the designer and executor use the same LLM backbone in our main experiments. 

\paragraph{Training Procedure.}
Our training proceeds in two stages: a supervised fine-tuning (SFT) cold start followed by reinforcement learning (RL) co-evolution. 
During the SFT stage, we initialize the policy by distilling trajectories from DeepSeek-V3.2 prompted with diverse workflow templates (further details regarding the cold start are provided in Appendix~\ref{sec:cold-start-details}). 
In the RL stage, we adopt stagewise designer-executor co-evolution with a stage length of $K=30$. For each query, the Designer generates $M=4$ candidate MAS, and each MAS is executed $N=4$ times. At each stage, only the active role is updated with a learning rate of $5\times 10^{-6}$, while gradients from the inactive role are masked.

\paragraph{Training Datasets.}
For the SFT cold start, the dataset consists of 3K Designer examples and 8K Executor examples, filtered from correct DeepSeek-V3.2 generations. 
For the RL stage, we train on a mixture of math and code data to encourage cross-task generalization. With an RL batch size of $8$, half of each batch is sampled from Polaris-Dataset-53K~\citep{Polaris2025}, and the remaining half is sampled from the APPS introductory subset~\citep{Hendrycks2021APPS} and CodeContests~\citep{DeepMind2024CodeContests}.
\paragraph{Baselines.}
We compare with four groups of baselines.
Single-agent baselines include direct prompting and GRPO, both using the same Qwen3-4B or 8B backbone as our method. GRPO is trained on the same math and code mixture.
Search-based MAS optimization baselines include AFlow~\citep{zhang2024aflow} and ADAS~\citep{hu2025automateddesignagenticsystems}. For AFlow, we use the official best-searched workflows for math and code. For ADAS, we use the official best-searched math agent and run the search protocol for code when no official code agent is released.
RL-based MAS optimization baselines include ScoreFlow~\citep{wang2025scoreflowmasteringllmagent}, MaAS~\citep{zhang2025multiagentarchitecturesearchagentic} and AFM~\citep{li2025chainofagents}.
For AFM~\citep{li2025chainofagents}, since the officially released checkpoint most comparable in scale to our setting is AFM-Coder-7B, we evaluate this checkpoint following the official code-agent evaluation framework.
All baselines follow the default settings in their original papers or released code. Details are given in Appendix~\ref{app:baseline}.
\paragraph{Benchmarks.}
We evaluate our models on both mathematical reasoning and code generation benchmarks. 
For math, we use \textsc{AIME24}/\textsc{AIME25}~\citep{AIME2024,AIME2025} and \textsc{OlympiadBench}~\citep{He2024OlympiadBench}. We evaluate each AIME benchmark 3 times and report the average.
All math tasks are evaluated with verifier-checked numeric scoring. 
For code, we use three widely adopted benchmarks: 
APPS~\citep{Hendrycks2021APPS}, 
LiveCodeBench-v6~\citep{jain2024livecodebench}, 
and CodeContests~\citep{DeepMind2024CodeContests}. 
Code tasks are evaluated by executing generated solutions against the official or benchmark-provided test cases.

\begin{table*}[t]
\centering
\caption{\textbf{Qwen3 8B} results on coding and math benchmarks. Parentheses denote gain over the Single Agent baseline. \colorbox{yellow!40}{Best} and \colorbox{gray!25}{second best} results are highlighted per benchmark.}
\label{tab:qwen3_8b}
\tiny
\setlength{\tabcolsep}{2pt}
\renewcommand{\arraystretch}{1}
\begin{tabular}{l| l ccc ccc c}
\toprule
& & \multicolumn{3}{c}{\textbf{Code}} & \multicolumn{3}{c}{\textbf{Math}} & \\
\cmidrule(lr){3-5} \cmidrule(lr){6-8}
\textbf{Training Paradigm} & \textbf{Method} 
& LiveCodeBench & APPS & CodeContests 
& AIME24 & AIME25 & OlympiadBench & \textbf{Avg} \\
\midrule

\multirow{2}{*}{\textbf{Single Agent}}
& SA
& \makecell{22.80\\(+0.00)}
& \makecell{30.20\\(+0.00)}
& \makecell{15.75\\(+0.00)}
& \makecell{18.30\\(+0.00)}
& \makecell{20.90\\(+0.00)}
& \makecell{55.00\\(+0.00)}
& \makecell{27.16\\(+0.00)} \\
& SA + GRPO
& \makecell{25.70\\(+2.90)}
& \cellcolor{gray!25}\makecell{37.00\\(+6.80)}
& \makecell{12.12\\(-3.63)}
& \makecell{18.30\\(+0.00)}
& \makecell{26.67\\(+5.77)}
& \makecell{54.80\\(-0.20)}
& \makecell{29.10\\(+1.94)} \\

\midrule
\multirow{2}{*}{\textbf{Search-based Auto MAS}}
& AFlow
& \makecell{28.60\\(+5.80)}
& \makecell{27.40\\(-2.80)}
& \makecell{15.80\\(+0.05)}
& \makecell{16.67\\(-1.63)}
& \makecell{20.83\\(-0.07)}
& \makecell{35.31\\(-19.69)}
& \makecell{24.10\\(-3.06)} \\
& ADAS
& \makecell{20.00\\(-2.80)}
& \makecell{27.00\\(-3.20)}
& \makecell{12.20\\(-3.55)}
& \makecell{13.30\\(-5.00)}
& \makecell{16.70\\(-4.20)}
& \makecell{32.90\\(-22.10)}
& \makecell{20.35\\(-6.81)} \\

\midrule
\multirow{5}{*}{\textbf{RL-based Auto MAS}}
& ScoreFlow
& \makecell{25.90\\(+3.10)}
& \makecell{26.50\\(-3.70)}
& \makecell{13.30\\(-2.45)}
& \makecell{28.90\\(+10.60)}
& \makecell{20.00\\(-0.90)}
& \makecell{51.30\\(-3.70)}
& \makecell{27.65\\(+0.49)} \\
& MaAS
& \makecell{24.29\\(+1.49)}
& \makecell{30.00\\(-0.20)}
& \makecell{15.15\\(-0.60)}
& \cellcolor{yellow!40}\makecell{45.80\\(+27.50)}
& \cellcolor{gray!25}\makecell{29.20\\(+8.30)}
& \makecell{48.90\\(-6.10)}
& \cellcolor{gray!25}\makecell{32.22\\(+5.06)} \\
& AFM-Coder
& \makecell{29.10\\(+6.30)}
& \makecell{28.00\\(-2.20)}
& \cellcolor{yellow!40}\makecell{21.20\\(+5.45)}
& \makecell{12.00\\(-6.30)}
& \makecell{8.00\\(-12.90)}
& \makecell{21.80\\(-33.20)}
& \makecell{20.35\\(-6.81)} \\

&\textbf{\name\textsubscript{SFT}}
& \cellcolor{gray!25}\makecell{\textbf{36.00}\\\textbf{(+13.20)}}
& \makecell{\textbf{32.00}\\\textbf{(+1.80)}}
& \makecell{\textbf{13.00}\\\textbf{(-2.75)}}
& \makecell{\textbf{33.00}\\\textbf{(+14.70)}}
& \makecell{\textbf{20.00}\\\textbf{(-0.90)}}
& \cellcolor{gray!25}\makecell{\textbf{59.00}\\\textbf{(+4.00)}}
& \makecell{\textbf{32.17}\\\textbf{(+5.01)}} \\
& \textbf{\name\textsubscript{RL}}
& \cellcolor{yellow!40}\makecell{\textbf{41.00}\\\textbf{(+18.20)}}
& \cellcolor{yellow!40}\makecell{\textbf{38.00}\\\textbf{(+7.80)}}
& \cellcolor{gray!25}\makecell{\textbf{17.00}\\\textbf{(+1.25)}}
& \cellcolor{gray!25}\makecell{\textbf{40.00}\\\textbf{(+21.70)}}
& \cellcolor{yellow!40}\makecell{\textbf{33.33}\\\textbf{(+12.10)}}
& \cellcolor{yellow!40}\makecell{\textbf{61.00}\\\textbf{(+6.00)}}
& \cellcolor{yellow!40}\makecell{\textbf{38.33}\\\textbf{(+11.17)}} \\

\bottomrule
\end{tabular}
\vspace{-10pt}
\end{table*}
\begin{table*}[h]
\centering
\caption{\textbf{Qwen3 4B} results on coding and math benchmarks. Parentheses denote gain over the Single Agent baseline. \colorbox{yellow!40}{Best} and \colorbox{gray!25}{second best} results are highlighted per benchmark.}
\label{tab:qwen3_4b}
\tiny
\setlength{\tabcolsep}{2pt}
\renewcommand{\arraystretch}{1}
\begin{tabular}{l| l ccc ccc c}
\toprule
& & \multicolumn{3}{c}{\textbf{Code}} & \multicolumn{3}{c}{\textbf{Math}} & \\
\cmidrule(lr){3-5} \cmidrule(lr){6-8}
\textbf{Training Paradigm} & \textbf{Method} 
& LiveCodeBench & APPS & CodeContests 
& AIME24 & AIME25 & OlympiadBench & \textbf{Avg} \\
\midrule

\multirow{2}{*}{\textbf{Single Agent}}
& SA
& \makecell{13.80\\(+0.00)}
& \makecell{27.40\\(+0.00)}
& \cellcolor{gray!25}\makecell{14.80\\(+0.00)}
& \makecell{20.00\\(+0.00)}
& \makecell{19.10\\(+0.00)}
& \makecell{33.20\\(+0.00)}
& \makecell{21.38\\(+0.00)} \\
& SA + GRPO
& \makecell{16.70\\(+2.90)}
& \cellcolor{gray!25}\makecell{35.60\\(+8.20)}
& \cellcolor{yellow!40}\makecell{18.60\\(+3.80)}
& \makecell{29.10\\(+9.10)}
& \cellcolor{yellow!40}\makecell{26.67\\(+7.57)}
& \makecell{47.10\\(+13.90)}
& \cellcolor{gray!25}\makecell{28.96\\(+7.58)} \\

\midrule
\multirow{2}{*}{\textbf{Search-based Auto MAS}}
& AFlow
& \makecell{28.00\\(+14.20)}
& \makecell{23.20\\(-4.20)}
& \makecell{13.33\\(-1.47)}
& \makecell{16.67\\(-3.33)}
& \makecell{13.33\\(-5.77)}
& \makecell{40.59\\(+7.39)}
& \makecell{22.52\\(+1.14)} \\
& ADAS
& \makecell{16.00\\(+2.20)}
& \makecell{28.00\\(+0.60)}
& \makecell{12.20\\(-2.60)}
& \makecell{10.00\\(-10.00)}
& \makecell{23.00\\(+3.90)}
& \makecell{32.80\\(-0.40)}
& \makecell{20.33\\(-1.05)} \\

\midrule
\multirow{4}{*}{\textbf{RL-based Auto MAS}}
& ScoreFlow
& \makecell{23.36\\(+9.56)}
& \makecell{24.50\\(-2.90)}
& \makecell{11.92\\(-2.88)}
& \makecell{26.40\\(+6.40)}
& \makecell{16.70\\(-2.40)}
& \cellcolor{gray!25}\makecell{57.00\\(+23.80)}
& \makecell{26.65\\(+5.27)} \\
& MaAS
& \makecell{24.29\\(+10.49)}
& \makecell{23.75\\(-3.65)}
& \makecell{9.10\\(-5.70)}
& \makecell{16.70\\(-3.30)}
& \cellcolor{gray!25}\makecell{25.00\\(+5.90)}
& \makecell{45.20\\(+12.00)}
& \makecell{24.01\\(+2.62)} \\

& \textbf{\name\textsubscript{SFT}}
& \cellcolor{gray!25}\makecell{\textbf{32.00}\\\textbf{(+18.20)}}
& \makecell{\textbf{32.00}\\\textbf{(+4.60)}}
& \makecell{\textbf{6.00}\\\textbf{(-8.80)}}
& \cellcolor{gray!25}\makecell{\textbf{30.00}\\\textbf{(+10.00)}}
& \makecell{\textbf{16.70}\\\textbf{(-2.40)}}
& \cellcolor{gray!25}\makecell{\textbf{57.00}\\\textbf{(+23.80)}}
& \makecell{\textbf{28.95}\\\textbf{(+7.57)}} \\
& \textbf{\name\textsubscript{RL}}
& \cellcolor{yellow!40}\makecell{\textbf{36.00}\\\textbf{(+22.20)}}
& \cellcolor{yellow!40}\makecell{\textbf{36.70}\\\textbf{(+9.30)}}
& \makecell{\textbf{14.20}\\\textbf{(-0.60)}}
& \cellcolor{yellow!40}\makecell{\textbf{33.33}\\\textbf{(+13.33)}}
& \cellcolor{yellow!40}\makecell{\textbf{26.67}\\\textbf{(+7.57)}}
& \cellcolor{yellow!40}\makecell{\textbf{58.20}\\\textbf{(+25.00)}}
& \cellcolor{yellow!40}\makecell{\textbf{34.18}\\\textbf{(+12.80)}} \\

\bottomrule
\end{tabular}
\vspace{-10pt}
\end{table*}

\subsection{Main Results}



Tables~\ref{tab:qwen3_8b} and~\ref{tab:qwen3_4b} report the performance of our cold-start and RL-trained models on six math and code benchmarks. Compared with the single-agent GRPO baseline, \name{}\textsubscript{RL} consistently achieves stronger performance across benchmarks. By introducing agent collaboration, the RL-based Auto MAS paradigm effectively overcomes the bottlenecks of isolated generation; for instance, \name\textsubscript{RL} reaches an impressive average accuracy of $38.33\%$ on Qwen3-8B and $34.18\%$ on Qwen3-4B, yielding absolute gains of $+11.17\%$ and $+12.80\%$ over the Single Agent baseline, respectively.

Search-based Auto MAS baselines generally perform poorly when instantiated with Qwen3-4B and Qwen3-8B. Methods like AFlow and ADAS frequently cause performance degradation  (e.g., ADAS drops by $-6.81\%$ to an average of $20.35\%$ on 8B, and AFlow only achieves $22.52\%$ on 4B). This significant drop indicates that search-based methods struggle to generalize across different model scales and are overly reliant on the underlying base models.

In contrast, \name\textsubscript{RL} transcends these limitations and surpasses all evaluated baselines. From \name\textsubscript{SFT} to \name\textsubscript{RL}, our model improves by 6.17\% on average, advancing from a suboptimal foundation to state-of-the-art average performance and demonstrating the effectiveness of our RL pipeline. Compared to methods that solely train a meta-agent or optimize workflow selection via RL (e.g., MaAS and ScoreFlow), \name\textsubscript{RL} successfully breaks the performance ceiling of static executors, outperforming the strong MaAS baseline by $+6.11\%$ on average ($38.33\%$ vs. $32.22\%$) on the 8B model. Furthermore, when compared to AFM-Coder, which shows severe performance imbalance and degrades heavily on math tasks, \name{} demonstrates exceptional cross-task generalization. Moreover, by explicitly adopting a multi-stage training paradigm, \name{} provides clearer and more targeted training signals, enabling a highly effective co-evolution of the agents' collaborative capabilities across diverse domains.




\subsection{Ablation Studies}
To isolate the factors driving stable and scalable meta-agent optimization, we ablate two central components of our method: Executor-Designer Hierarchical Rollout and Stagewise Co-evolution. We also study the architectural design space by comparing shared- and separate-policy training.

\paragraph{Executor-Designer Hierarchical Rollout.}

\begin{wrapfigure}{r}{0.4\linewidth}
\vspace{-1.2em}
\centering
\captionsetup{font=scriptsize,skip=2pt}
\captionof{table}{Ablation results of Executor-Designer Hierarchical Rollout settings.}
\label{tab:rollout_var}
\footnotesize
\setlength{\tabcolsep}{4.5pt}
\renewcommand{\arraystretch}{1.05}
\begin{tabular}{lcc}
\toprule
\textbf{Rollout} & \textbf{AIME24} & \textbf{AIME25} \\
\midrule
$M{=}4,N{=}4$ & 40.0\% & 33.3\% \\
$M{=}8,N{=}1$ & 33.3\% & 30.0\% \\
\bottomrule
\end{tabular}
\vspace{-1.0em}
\end{wrapfigure}

We compare different hierarchical rollout configurations. Our main experiments use $M{=}4$ and $N{=}4$, where each query samples four candidate designs and executes each design four times. We additionally evaluate a flatter rollout setting with $M{=}8$ and $N{=}1$, where more designs are sampled but each design is executed only once. As shown in Table~\ref{tab:rollout_var}, the hierarchical setting achieves better performance, improving AIME24 from 33.3\% to 40.0\% and AIME25 from 30.0\% to 33.3\%. This suggests that repeated executions of each sampled design provide a more reliable estimate of downstream utility, leading to more stable credit assignment.

\paragraph{Does Stagewise Co-evolution Help?}

\begin{wrapfigure}{r}{0.45\linewidth}
\vspace{-1.3em}
\centering
\includegraphics[width=\linewidth]{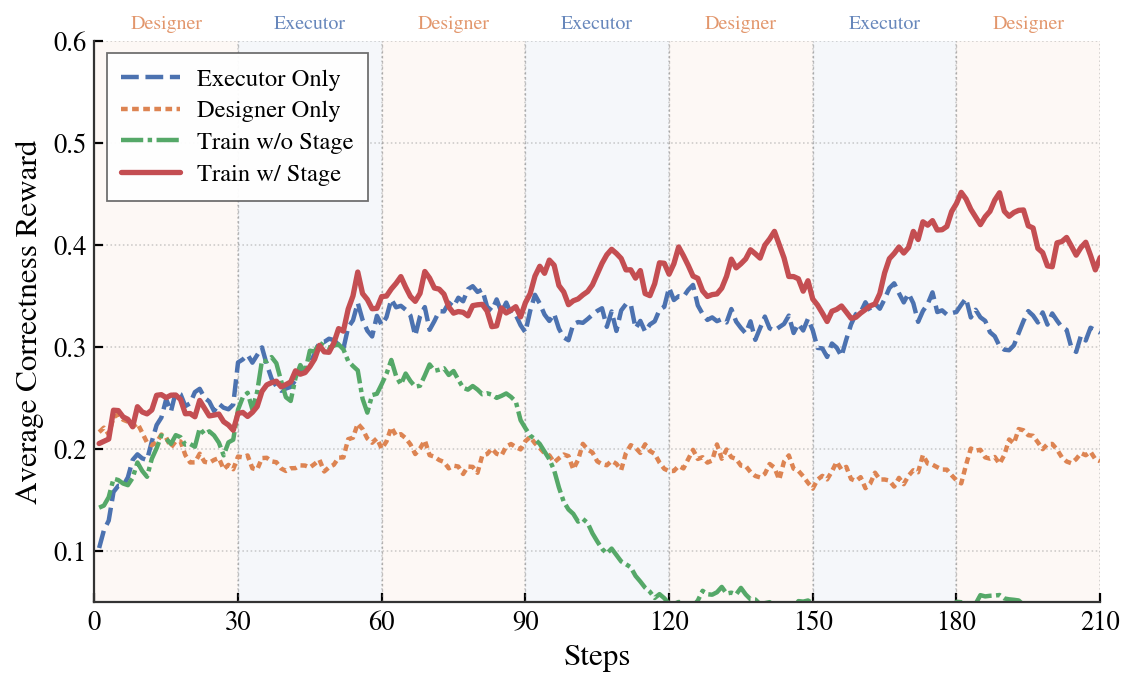}
\vspace{-2em}
\caption{Training-reward dynamics ablations of the proposed stagewise co-evolution.}
\label{fig:ablation-reward}
\vspace{-1em}
\end{wrapfigure}

We compare the proposed schedule on Qwen3-8B with three variants: coupled training, executor-only training, and designer-only training. In the coupled setting, trajectories from both roles update the shared policy simultaneously. As shown in Figure~\ref{fig:ablation-reward}, this variant improves quickly at first but later collapses; during evaluation, the model often repeats meaningless tokens until reaching the maximum length. Designer-only training brings limited improvement, suggesting that optimizing designs alone does not reliably improve MAS performance. Executor-only training improves correctness rapidly but soon saturates, indicating a ceiling imposed by the fixed design policy.
\begin{wraptable}[7]{r}{0.35\columnwidth}
\vspace{-1.0em}
\centering
\captionsetup{font=scriptsize,skip=2pt}
\caption{Stagewise ablation.}
\label{tab:stagewise_ablation}
\scriptsize
\setlength{\tabcolsep}{3pt}
\renewcommand{\arraystretch}{1.03}
\begin{tabular*}{\linewidth}{@{\extracolsep{\fill}}lcc@{}}
\toprule
\textbf{Variant} & \textbf{Math} & \textbf{Code} \\
\midrule
Coupled       & 36.7\% & 25.2\% \\
Designer-only & 38.6\% & 27.5\% \\
Executor-only & 39.6\% & 30.7\% \\
Stagewise     & 44.8\% & 32.0\% \\
\bottomrule
\end{tabular*}
\vspace{-1.0em}
\end{wraptable}

In contrast, stagewise training shows a clear staircase-shaped learning curve: reward remains relatively stable during designer phases and rises sharply after switching to executor phases. The accompanying table further shows that stagewise training achieves the best performance on both math and code benchmarks. These results suggest that stagewise Designer-Executor optimization provides a more stable and effective training path.

\paragraph{Shared Policy vs. Separate Policy.}
\begin{wraptable}[5]{r}{0.4\linewidth} \vspace{-1.2em} 
\centering 
\captionsetup{font=scriptsize,skip=2pt}
\captionof{table}{Ablation results of shared vs. separate policy.} \label{tab:policy_var} \footnotesize \setlength{\tabcolsep}{4.5pt} \renewcommand{\arraystretch}{1.05} \begin{tabular}{lcc} \toprule \textbf{Variant} & \textbf{AIME24} & \textbf{AIME25} \\ \midrule Shared & 40.0\% & 33.3\% \\ Separate & 33.3\% & 26.7\% \\ \bottomrule \end{tabular} \vspace{-1.0em} \end{wraptable}
We compare two policy parameterizations: a shared policy for both Designer and Executor, with role-specific prompts specifying their behaviors, and separate role-specific policies. As shown in Table~\ref{tab:policy_var}, the shared policy consistently outperforms separate policies on both AIME24 and AIME25. This suggests that Designer and Executor learning are not independent subtasks, but coupled components of the same meta-agent optimization problem. Sharing representations allows the training signal from one role to serve as an inductive bias for the other, improving generalization and data efficiency while reducing overfitting to role-specific trajectories.
\subsection{Analysis}
\label{sec:analysis}
We provide a more detailed analysis for the experiment results in Appendix~\ref{sec:appendix-case-studies}, including
stage-length sensitivity, per-query design diversity, and end-to-end case studies. Here we
summarize two main observations: RL changes both the \emph{structures emitted by the
Designer} and the \emph{quality of Executor behavior}.
\paragraph{Per-task structure selection.}

\begin{wraptable}[10]{r}{0.46\columnwidth}
\centering
\captionsetup{font=scriptsize,skip=2pt}
\caption{Structure share selected by the RL designer.}
\label{tab:rl-pattern-share}
\scriptsize
\setlength{\tabcolsep}{3pt}
\renewcommand{\arraystretch}{1.04}
\begin{tabular*}{\linewidth}{@{\extracolsep{\fill}}lccc@{}}
\toprule
\textbf{Benchmark} & \textbf{Single} & \textbf{Reflection} & \textbf{Ensemble} \\
\midrule
AIME 2024        & 18.9\% & \textbf{70.0\%} & 11.1\% \\
AIME 2025        & 15.6\% & \textbf{73.3\%} & 11.1\% \\
OlympiadBench    & \textbf{46.4\%} & 44.8\% & \phantom{0}8.8\% \\
\addlinespace[2pt]
CodeContests     & 26.7\% & \textbf{62.4\%} & 10.9\% \\
LiveCodeBench    & 43.5\% & \textbf{52.6\%} & \phantom{0}3.8\% \\
APPS             & \textbf{55.2\%} & 43.8\% & \phantom{0}1.0\% \\
\bottomrule
\end{tabular*}
\vspace{-0.8em}
\end{wraptable}

Table~\ref{tab:rl-pattern-share} reports the top three structures most frequently generated by the RL-trained designer. The \textbf{Single} structure uses one agent to solve the problem directly. The \textbf{Reflection} structure uses one agent to generate an initial solution and another agent to provide refinements. The \textbf{Ensemble} structure runs agents with different roles in parallel and uses a judge agent to select or synthesize the final answer. The structure selection is clearly task-dependent. On harder math benchmarks such as AIME, the designer selects \textbf{reflection} for more than 70\% of problems, indicating a preference for iterative verification on challenging reasoning tasks. On relatively easier tasks such as OlympiadBench and APPS, it routes a larger fraction of problems to a \textbf{single} structure. \textbf{Ensemble} is mainly selected for competition-style math and code tasks. These results suggest that MetaAgent learns to adapt the agent structure according to task characteristics.

\begin{wraptable}{r}{0.5\linewidth}
\vspace{-1.0em}
\centering
\tiny
\newcommand{\badcase}{\textcolor{red!75!black}{\textbf{\(\times\)}}}
\newcommand{\goodcase}{\textcolor{green!45!black}{\textbf{\(\checkmark\)}}}
\newcommand{\tagdesign}{\textcolor{blue!70!black}{\textbf{Better Design}}}
\newcommand{\tagexec}{\textcolor{orange!85!black}{\textbf{Better Execute}}}
\setlength{\tabcolsep}{2.5pt}
\renewcommand{\arraystretch}{1.05}
\begin{tabular}{p{0.25\linewidth}p{0.36\linewidth}p{0.36\linewidth}}
\toprule
\textbf{Example} & \textbf{SFT Model} & \textbf{RL Model} \\
\midrule
Math\\
\tagdesign &
\badcase\ Ensemble judge. All solvers share the wrong circle-packing model;
judge reports contradiction but cannot repair. &
\goodcase\ Solver--critic reflection. Critic localizes the geometry error;
solver switches to similar triangles. \\
\midrule
Code\\
\tagexec &
\badcase\ Same reflection structure, but executor keeps double-counting
divisors after sample outputs are $2\times$ too large. &
\goodcase\ Same reflection structure, but executor uses tests to restore the
one-count-per-divisor invariant. \\
\bottomrule
\end{tabular}
\caption{SFT-to-RL case comparison: RL improves both MAS design and executor
repair behavior.}
\label{tab:sft-rl-case-study}
\vspace{-1.0em}
\end{wraptable}
\paragraph{Which role brings the improvement?}
To disentangle whether the designer or the executor is the primary driver of this success, we analyzed AIME25, and the results show: half of the improvements stem from the executor successfully solving the problem under the \emph{same} structural pattern assigned by SFT, demonstrating clear execution-side capability gains. The remaining $50\%$ of improvements occur when the designer flips to a more efficient pattern. We include two qualitative examples to illustrate how RL changes model
behavior after cold start. The math example highlights better \emph{design}
(choosing a repairable MAS structure), while the code example highlights better
\emph{execution} under the same solver--tester structure. Details are in Appendix~\ref{sec:appendix-case-studies}.

\section{Discussions}
We introduced \name, the first end-to-end reinforcement learning framework that jointly optimizes the designer and executor of an automatic multi-agent system through hierarchical rollouts and stagewise co-evolution. Across six math and code benchmarks and two model scales, \name consistently surpasses both human-designed and existing automatic MAS baselines by up to 21.7\%, while exposing the internal dynamics through which designer and executor mutually improve. Moreover, \name suggests a path toward foundation models with native multi-agent capabilities, where MAS becomes an internal mechanism for reasoning and context management rather than an external, human-designed harness. However, our experiments are constrained by the computational resources, so we do not perform an exhaustive scaling study over larger backbone models or longer training budgets. Future work can further examine how the proposed trainable automatic MAS framework scales with model size, task diversity, and rollout budget.
\bibliographystyle{plainnat}
\bibliography{example_paper}
\newpage
\appendix
\section*{Appendix} 
\section{Cold Start Details}
\label{sec:cold-start-details}

We cold-start the policy by distilling both sides of the generated
multi-agent system: the \emph{Designer}, which writes an executable workflow,
and the \emph{Executors}, which solve the problem inside the generated
workflow. The Designer is prompted with a bank of workflow templates
implemented in our codebase. The template bank contains \textbf{single-agent}, \textbf{ensemble-voting}, \textbf{solver-critic reflection}, \textbf{solver-tester}, and etc. Each
template specifies both the workflow topology and the role-level prompts; for
example, ensemble templates instantiate strategy-diverse solvers and a judge,
while reflection templates instantiate a solver and a critic/verifier loop.

For each training question, we sample in-context examples from this template
bank and ask DeepSeek-V3.2 to synthesize a complete workflow program. The
data-generation pipeline samples multiple workflow designs per question and
logs both the Designer conversation and all Executor conversations produced
when the workflow is run. We then retain trajectories whose final answer is
judged correct, yielding 3K Designer examples and 8K Executor
examples for supervised cold start. This gives the model an initial ability
to map a problem to an appropriate multi-agent program and to act as the
specialized agents inside that program.

Before training our smaller policy, we also evaluate whether the same
workflow prompting strategy helps a strong proprietary model. As shown in
Table~\ref{tab:cold-start-deepseek-prompting}, prompting DeepSeek to solve
through a generated multi-agent workflow improves AIME 2024 accuracy from
63.3\% to 66.7\%, which suggests that the workflow
interface is not only a crutch for weaker models: even when the underlying
model is already strong, explicit role decomposition and verification can
recover additional correct solutions. This motivates using DeepSeek-V3.2 as
the teacher for cold-starting both the Designer and Executor behaviors before
reinforcement learning.

\begin{table}[h]
\centering
\small
\setlength{\tabcolsep}{8pt}
\begin{tabular}{lcc}
\toprule
Method & Accuracy \\
\midrule
Direct DeepSeek & 63.3\% \\
DeepSeek prompted with MAS workflow &  66.7\% \\
\bottomrule
\end{tabular}
\caption{Prompting-only comparison on AIME 2024 using DeepSeek as the
underlying model. The MAS prompt asks DeepSeek to first synthesize and run a
multi-agent workflow rather than directly answer with a single response.}
\label{tab:cold-start-deepseek-prompting}
\end{table}

\section{Reward Design}
\label{sec:reward-design}

The outcome-based reward $R(e_{i,j}, d_i)$ is composed of two terms:
\begin{equation}
    R(e_{i,j}, d_i) = R_{\text{correct}}(e_{i,j}) + \lambda \cdot R_{\text{format}}(e_{i,j}),
    \label{eq:reward}
\end{equation}
where $R_{\text{correct}}$ evaluates the functional correctness of the final solution via environment feedback, and $R_{\text{format}}$ incentivizes structured agent behavior. We set $\lambda=0.4$.

The correctness reward $R_{\text{correct}} \in \{0, 1\}$ is a strict binary signal determined by the specific domain environment:
\begin{itemize}[leftmargin=*]
    \item \textbf{Math Verification:} The final parsed answer is evaluated against the ground-truth solution. To account for algebraically equivalent expressions, we utilize a symbolic math engine (e.g., SymPy) to robustly verify the correctness of the final mathematical output.
    \item \textbf{Code Execution:} The final generated program is compiled and executed against the dataset's hidden unit tests. The reward $R_{\text{correct}} = 1$ is assigned if and only if the code successfully passes all unit tests without exceeding the environment's execution time or memory constraints.
\end{itemize}

The format reward $R_{\text{format}}$ acts as a regularizer and consists of two components:
\begin{enumerate}[leftmargin=*]
    \item \textbf{Solution formatting.} The final agent must produce its answer within a standardized output format, ensuring that the solution is reliably parseable for automated evaluation.
    \item \textbf{Delivery formatting.} Inter-agent messages must be strictly enclosed within \texttt{<delivery>}...\texttt{</delivery>} tags. This constraint serves a dual purpose: it establishes a structured, easily parsable communication protocol, and crucially, it incentivizes agents to \emph{distill} relevant information into concise deliverables rather than forwarding their entire reasoning trace. Without this constraint, agents tend to broadcast full outputs, unnecessarily inflating the context window without improving coordination quality.
\end{enumerate}
\section{Result Analysis
}
\begin{figure}[t]
    \centering
    \includegraphics[width=1\linewidth]{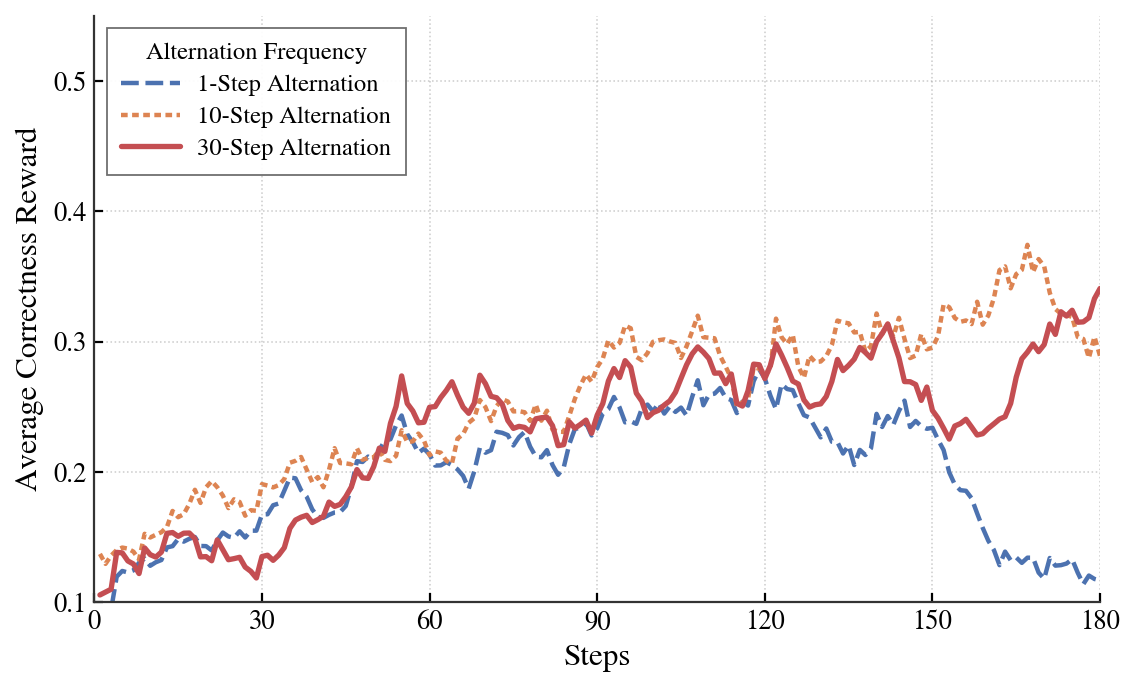}
    \caption{
    Sensitivity analysis on the stage length for designer--executor alternation.
    One-step alternation is unstable and collapses during training, while longer stages provide more stable role-specific optimization. The 30-step schedule achieves the best final reward and is used as the default setting in our main experiments.
    }
    \label{fig:stage_length_ablation}
\end{figure}

\paragraph{Sensitivity Analysis on Stage Length.}
Given that stagewise optimization is essential for stable designer--executor co-evolution, we further ablate how frequently the active role should be switched. Figure~\ref{fig:stage_length_ablation} compares three alternation intervals: 1-step, 10-step, and 30-step switching. Alternating the active role at every step leads to highly unstable training: neither role accumulates sufficient role-consistent gradient signal before being interrupted, and the training run collapses after approximately 150 steps. Increasing the interval to 10 or 30 steps substantially improves stability. Among them, the 30-step schedule achieves the highest final reward and exhibits the clearest upward trend. We therefore adopt 30-step alternation as the default setting in all main experiments.

\paragraph{Per-query design diversity.}

Besides the structures, MetaAgent design different task-specific and various agents role. Across all the workflows the designer emits in evaluation, it
instantiates a vocabulary of \textbf{54 distinct role names}; System prompts are even more diverse:
\textbf{77.5\% of agents receive a byte-unique, task-specific prompt}. Each workflow is therefore a freshly synthesised program
tailored to the problem at hand, not an instantiation of a pre-defined
template; the structural taxonomy in
Table~\ref{tab:rl-pattern-share} is a coarse skeleton, while the body
of every workflow is per-query content.

\paragraph{Which role brings the improvement?}
RL changes \emph{both} (i)~which structure the designer selects and
(ii)~how well the executor performs under it. The designer's routing
distribution moves uniformly across math benchmarks
(Table~\ref{tab:sft-rl-shift}): \textbf{ensemble+judge} loses
$\sim\!30$~pp on every benchmark and the share is redirected mostly to
\textbf{reflection}, with a smaller push toward \textbf{single}. To
disentangle the two effects we look at the same 30 AIME 2025 problems
solved by SFT and RL. Of the 
problems RL solves but SFT does not, 50\% use the \emph{same} pattern as
SFT, which show an executor-side improvement. And 50\% are produced by a designer
flip to a different (and in those cases simpler) pattern, indicating the improvement from
SFT$\to$RL also benefits from a designer flip.

\begin{table}[h]
\centering
\begin{minipage}[c]{0.46\linewidth}
\centering\small
\setlength{\tabcolsep}{8pt}
\begin{tabular}{@{}ll@{}}
\toprule
\multicolumn{2}{c}{Top-10 role names} \\
\midrule
\textsf{CodeSolver}      & \textsf{AlgebraicSolver}     \\
\textsf{MathSolver}      & \textsf{BruteForceSolver}    \\
\textsf{UnitTestAgent}   & \textsf{CombinatorialSolver} \\
\textsf{MathCritic}      & \textsf{EdgeCaseSolver}      \\
\textsf{MathJudge}       & \textsf{OptimalSolver}       \\
\bottomrule
\end{tabular}
\captionof{table}{Top-10 role names the designer emits across
$2{,}574$ workflows.}
\label{tab:top-roles}
\end{minipage}\hfill
\begin{minipage}[c]{0.50\linewidth}
\centering\small
\setlength{\tabcolsep}{3pt}
\begin{tabular}{lcccccc}
\toprule
& \multicolumn{2}{c}{\textbf{single}} & \multicolumn{2}{c}{\textbf{refl.}} & \multicolumn{2}{c}{\textbf{ens.}} \\
\cmidrule(lr){2-3}\cmidrule(lr){4-5}\cmidrule(lr){6-7}
Benchmark      & SFT  & RL   & SFT  & RL   & SFT  & RL   \\
\midrule
AIME 2024      & 16.7 & 18.9 & 38.9 & 70.0 & 44.4 & 11.1 \\
AIME 2025      & \phantom{0}7.8 & 15.6 & 41.1 & 73.3 & 51.1 & 11.1 \\
OlympiadBench  & 42.3 & 46.4 & 21.1 & 44.8 & 36.6 & \phantom{0}8.8 \\
\bottomrule
\end{tabular}
\captionof{table}{Pattern share (\%) on math benchmarks: SFT cold-start
vs.\ RL. }
\label{tab:sft-rl-shift}
\end{minipage}
\end{table}

\section{Baseline Details}
\label{app:baseline}

We describe the baseline implementations used in our experiments. Unless otherwise stated, all baselines use the same execution backbone as our method, instantiated with Qwen3 4B or Qwen3 8B according to the corresponding experimental setting. All reported results are evaluated on the same final test split and metric as our method. For baselines that require search or training, we follow the default protocol of the original paper or released code, while matching the rollout budget of our method whenever the method exposes the corresponding parameter.

\subsection{Search Based MAS Optimization Baselines}

\paragraph{AFlow.}
AFlow~\citep{zhang2024aflow} searches over code represented agentic workflows with MCTS. We follow the official paper and code settings. Specifically, AFlow uses \texttt{sample}=4, \texttt{initial\_round}=1, \texttt{max\_rounds}=20, \texttt{validation\_rounds}=5, and early stopping enabled. The \texttt{sample}=4 setting matches the number of candidate MAS designs generated by our Designer for each query. During workflow evaluation, we execute each candidate workflow $N=4$ times when stochastic execution is supported, matching our execution budget. We keep the original domain specific operator sets: \texttt{Custom}, \texttt{ScEnsemble}, and \texttt{Programmer} for math, and \texttt{Custom}, \texttt{CustomCodeGenerate}, \texttt{ScEnsemble}, and \texttt{Test} for code. For each task domain, we use the best searched workflow reported or released by AFlow and evaluate it on our held out test split.

\paragraph{ADAS.}
ADAS~\citep{hu2025automateddesignagenticsystems} performs Meta Agent Search, where a meta agent writes executable Python \texttt{forward} functions and adds evaluated agents to an archive. We follow the official split and search protocol. For MGSM and related reasoning domains, the official implementation shuffles all examples with seed 0, uses 128 examples for search validation, and uses the next 800 examples for final testing. For GPQA diamond, it uses 32 validation examples and the remaining 166 examples for testing. For ARC, it uses 20 validation tasks and 60 test tasks, with five repeated evaluations to reduce stochastic variance.

The default MGSM search uses \texttt{n\_generation}=30, \texttt{n\_repeat}=1, \texttt{max\_workers}=48, and at most three debugging attempts for invalid generated code. To align with our training budget, when we run ADAS search on a new benchmark without an official task specific searched agent, each generation evaluates $M=4$ newly proposed candidate agents when supported by the search implementation, and each candidate agent is executed $N=4$ times for reward estimation. The candidate agent executor follows the same Qwen3 4B or Qwen3 8B backbone as our method. The reflection call uses temperature 0.8. The initial archive follows the released implementation and contains self consistency with Chain of Thought, Self Refine, LLM Debate, Step Back Abstraction, Quality Diversity, and Role Assignment. For math, we use the best searched MGSM agent reported by ADAS, \textbf{Dynamic Role Playing Architecture}, and keep its original role routing and answer aggregation. For scode, the official ADAS repository does not release a task specific searched code agent. Therefore, when no official task specific searched code agent is available, we run Meta Agent Search on the corresponding optimization split with the same default ADAS search budget and select the best validation agent for held-out test evaluation.

\subsection{Semi Learning Based MAS Optimization Baselines}

\paragraph{ScoreFlow.}
ScoreFlow~\citep{wang2025scoreflowmasteringllmagent} trains a workflow generator with Score DPO. We follow the released training and inference pipeline, while aligning the candidate and execution budget with our setting. For each optimization step, ScoreFlow generates $M=4$ candidate workflows per query and executes each candidate workflow $N=4$ times to estimate its score. The resulting workflow scores are used to construct preference pairs for Score DPO. The held out test split is used only for final inference. For inference, we use the trained checkpoint selected by the original validation protocol.

We use the same stage length and learning rate as our method whenever ScoreFlow updates trainable parameters. Specifically, optimization is organized into stages of length $K=30$, and the workflow generator is updated with learning rate $5\times 10^{-6}$. Only the trainable generator parameters are updated; executor parameters are frozen. Score DPO is implemented with LoRA, with rank 8, alpha 16, dropout 0.01, target modules \texttt{q\_proj} and \texttt{v\_proj}, and no bias. The generator uses temperature 0.2, top $p=0.95$, and maximum generation length 1000. The executor uses temperature 0.0. The vLLM setting uses bfloat16, GPU memory utilization 0.9, and maximum model length 10000. Each optimization epoch trains for 1 epoch with per device train batch size 1 and per device evaluation batch size 1. Logging is performed every 10 steps, and LoRA weights are merged into the generator checkpoint after each epoch.

\paragraph{MaAS.}
MaAS~\citep{zhang2025multiagentarchitecturesearchagentic} trains an agentic supernet and samples query dependent architectures from it. We follow the official training and evaluation protocol. Each benchmark is split into train and test sets with a $1:4$ ratio. The training split is used to optimize the controller distribution and agentic operators, while the held out test split is used only for final evaluation. We match the architecture sampling and training budget to our method where possible. For each query, MaAS samples $M=4$ candidate architectures from the agentic supernet and executes each sampled architecture $N=4$ times for reward estimation. We use a stage length of $K=30$ and update the trainable controller and architecture parameters with learning rate $5\times 10^{-6}$. The execution backbone remains fixed. We keep the official architectural defaults: maximum supernet depth $L=4$, sampling times $K=4$ in the original MaAS notation, early exit threshold 0.3, and cost penalty coefficient selected from $\{1e{-}3, 5e{-}3, 1e{-}2\}$. In the released command, optimization is run with \texttt{sample}=4; the same command is then rerun with \texttt{is\_test=True} for held out evaluation. We use the best validation setting selected by the original protocol and report its held out test performance.

\paragraph{Agent Foundation Model.}
Agent Foundation Model (AFM)~\citep{li2025chainofagents} is an end-to-end agent model trained under the Chain-of-Agents paradigm. Instead of explicitly instantiating an external multi-agent workflow at test time, AFM internalizes multi-agent collaboration into a single model through multi-agent distillation and agentic reinforcement learning. This makes AFM different from our domain-adaptive automatic MAS setting: AFM is a released agent foundation model trained with its own data and backbone, while our method learns to construct and execute task-specific MAS using the Qwen3 4B or Qwen3 8B backbone. Nevertheless, AFM is a relevant baseline because it represents a strong end-to-end alternative to explicit MAS optimization.

We evaluate the officially released size-comparable code-agent checkpoint, \texttt{AFM-CodeAgent-7B-rl}. The released AFM model card also lists larger \texttt{AFM-CodeAgent-32B} checkpoints, but we use the 7B checkpoint to keep the comparison closer to our 8B experimental scale. We follow the official AFM code-agent evaluation framework. We use the official code-agent evaluation script with the released default parameters. The maximum prompt length is set to $4096$ tokens, and the maximum response length is set to $28672$ tokens. The rollout uses \texttt{n}=8 samples, \texttt{val\_kwargs.temperature}=0.6, and multi-turn tool use with at most 12 turns. 

%

\definecolor{cs-question}{HTML}{2D3748}
\definecolor{cs-question-bg}{HTML}{F7FAFC}
\definecolor{cs-designer}{HTML}{2C5282}
\definecolor{cs-designer-bg}{HTML}{EBF4FB}
\definecolor{cs-agent}{HTML}{276749}
\definecolor{cs-agent-bg}{HTML}{F0FFF4}
\definecolor{cs-judge}{HTML}{B7791F}
\definecolor{cs-judge-bg}{HTML}{FFFBEB}
\definecolor{cs-answer}{HTML}{9B2C2C}
\definecolor{cs-answer-bg}{HTML}{FFF5F5}
\definecolor{cs-elide}{HTML}{718096}
\definecolor{cs-tool-bg}{HTML}{2D3748}
\definecolor{cs-tool-fg}{HTML}{E2E8F0}
\definecolor{cs-code-bg}{HTML}{1A202C}
\definecolor{cs-code-fg}{HTML}{E2E8F0}
\definecolor{cs-code-kw}{HTML}{F6AD55}
\definecolor{cs-code-str}{HTML}{68D391}
\definecolor{cs-code-com}{HTML}{A0AEC0}

\newcommand{\cmark}{\textcolor{cs-agent}{\ding{51}}}
\newcommand{\xmark}{\textcolor{cs-answer}{\ding{55}}}

\lstdefinestyle{cspython}{
  language=Python,
  basicstyle=\ttfamily\scriptsize\color{cs-code-fg},
  backgroundcolor=\color{cs-code-bg},
  keywordstyle=\color{cs-code-kw}\bfseries,
  stringstyle=\color{cs-code-str},
  commentstyle=\color{cs-code-com}\itshape,
  showstringspaces=false,
  breaklines=true,
  breakatwhitespace=true,
  columns=fullflexible,
  frame=none,
  framesep=4pt,
  xleftmargin=8pt,
  xrightmargin=4pt,
  aboveskip=2pt,belowskip=2pt,
}
\lstdefinestyle{cstool}{
  basicstyle=\ttfamily\scriptsize\color{cs-tool-fg},
  backgroundcolor=\color{cs-tool-bg},
  showstringspaces=false,
  breaklines=true,
  columns=fullflexible,
  frame=none,
  framesep=4pt,
  xleftmargin=8pt,
  xrightmargin=4pt,
  aboveskip=2pt,belowskip=2pt,
}

\newtcolorbox{questionbox}[1][]{
  enhanced, breakable, colback=cs-question-bg, colframe=cs-question,
  boxrule=0.4pt, arc=2pt, left=8pt, right=8pt, top=5pt, bottom=5pt,
  fontupper=\small,
  title=\textbf{Question}, fonttitle=\bfseries\footnotesize\color{white},
  coltitle=white, colbacktitle=cs-question, attach boxed title to top left={yshift=-2mm,xshift=4mm},
  boxed title style={colback=cs-question, sharp corners, boxrule=0pt},
  #1
}

\newtcolorbox{designerbox}[1][]{
  enhanced, breakable, colback=cs-designer-bg, colframe=cs-designer,
  boxrule=0.4pt, arc=2pt, left=8pt, right=8pt, top=5pt, bottom=5pt,
  fontupper=\footnotesize,
  title=\textbf{Designer (raw output)}, fonttitle=\bfseries\footnotesize\color{white},
  coltitle=white, colbacktitle=cs-designer, attach boxed title to top left={yshift=-2mm,xshift=4mm},
  boxed title style={colback=cs-designer, sharp corners, boxrule=0pt},
  #1
}

\newtcolorbox{agentbox}[2][]{
  enhanced, breakable, colback=cs-agent-bg, colframe=cs-agent,
  boxrule=0.4pt, arc=2pt, left=8pt, right=8pt, top=5pt, bottom=5pt,
  fontupper=\footnotesize,
  title=\textbf{Executor: #2}, fonttitle=\bfseries\footnotesize\color{white},
  coltitle=white, colbacktitle=cs-agent, attach boxed title to top left={yshift=-2mm,xshift=4mm},
  boxed title style={colback=cs-agent, sharp corners, boxrule=0pt},
  #1
}

\newtcolorbox{judgebox}[1][]{
  enhanced, breakable, colback=cs-judge-bg, colframe=cs-judge,
  boxrule=0.4pt, arc=2pt, left=8pt, right=8pt, top=5pt, bottom=5pt,
  fontupper=\footnotesize,
  title=\textbf{Judge}, fonttitle=\bfseries\footnotesize\color{white},
  coltitle=white, colbacktitle=cs-judge, attach boxed title to top left={yshift=-2mm,xshift=4mm},
  boxed title style={colback=cs-judge, sharp corners, boxrule=0pt},
  #1
}

\newtcolorbox{toolbox}[1][]{
  enhanced, breakable, colback=cs-tool-bg, colframe=cs-tool-bg,
  boxrule=0pt, arc=2pt, left=8pt, right=8pt, top=4pt, bottom=4pt,
  fontupper=\ttfamily\scriptsize\color{cs-tool-fg},
  #1
}

\newtcolorbox{answerbox}[1][]{
  enhanced, breakable, colback=cs-answer-bg, colframe=cs-answer,
  boxrule=0.4pt, arc=2pt, left=8pt, right=8pt, top=5pt, bottom=5pt,
  fontupper=\small,
  title=\textbf{Final Answer}, fonttitle=\bfseries\footnotesize\color{white},
  coltitle=white, colbacktitle=cs-answer, attach boxed title to top left={yshift=-2mm,xshift=4mm},
  boxed title style={colback=cs-answer, sharp corners, boxrule=0pt},
  #1
}

\newcommand{\elide}[1][\textit{(reasoning continues; not shown)}]{%
  \par\smallskip\noindent
  \textcolor{cs-elide}{\hspace*{1em}\ldots\quad{\scriptsize #1}\quad\ldots}\par\smallskip
}

\newcommand{\casestudy}[3]{%
  \par\medskip
  {\noindent\bfseries Case~#1: #2}\par
  \noindent\textcolor{cs-elide}{\footnotesize Domain: \textbf{#3}}\par\smallskip
}


\section{Case Studies}
\label{sec:appendix-case-studies}
\subsection{RL model Examples}
{\small
We present three end-to-end trajectories of our system, illustrating how the
\emph{designer} chooses a team structure for each question and how the
selected executors collaborate to reach the final answer. Cases span both
math and code domains and showcase three distinct team structures:
\textbf{single agent}, \textbf{ensemble + judge}, and \textbf{reflection
with a separate critic}. The designer and executor boxes contain
\emph{verbatim} model output. Only chain-of-thought stretches that do not
affect the exposition are elided, marked by the literal symbol
``$\ldots$\,\textit{(...)}\,$\ldots$''.
\par}

\casestudy{1}{Probability via single-agent reasoning}{Math (AIME 2024 \#2)}

\begin{questionbox}
Jen enters a lottery by picking $4$ distinct numbers from
$S=\{1,2,\dots,10\}$. $4$ numbers are randomly chosen from $S$. She wins a
prize if at least two of her numbers were among the $4$ chosen, and the
\emph{grand prize} if all four match. The probability of winning the grand
prize given that she won a prize is $\tfrac{m}{n}$ in lowest terms.
Find $m+n$.
\end{questionbox}

\begin{designerbox}
\textbf{Problem Type:} MATH

\textbf{Problem Analysis:}\\
This is a probability problem involving conditional probability.
\begin{itemize}[leftmargin=*,topsep=1pt,itemsep=0pt]
\item Jen picks 4 distinct numbers from $S = \{1,2,\dots,10\}$.
\item 4 numbers are randomly chosen from $S$ (presumably without replacement, since they are distinct).
\item She wins a prize if at least 2 of her numbers match the 4 chosen numbers.
\item She wins the grand prize if all 4 of her numbers match the 4 chosen numbers.
\item We are asked: given that she won a prize, what is the probability she won the grand prize?
\item The answer is $\frac{m}{n}$ in lowest terms, and we need $m+n$.
\end{itemize}

Mathematically:
Let $A$ = event she wins the grand prize (all 4 match).
Let $B$ = event she wins a prize (at least 2 match).
We need $P(A \mid B) = \frac{P(A)}{P(B)}$.

We can compute:
\begin{itemize}[leftmargin=*,topsep=1pt,itemsep=0pt]
\item Total ways to choose 4 numbers from 10: $\binom{10}{4}$.
\item If Jen's numbers are fixed (say $\{a,b,c,d\}$), then:
  $P(A)$ = probability the 4 chosen numbers are exactly Jen's 4 numbers: $\frac{1}{\binom{10}{4}}$.
  $P(B)$ = probability at least 2 of the 4 chosen numbers are in Jen's set.
  This can be computed as $1 - P(\text{0 or 1 match})$.
\end{itemize}

We can also think:
Given that she won a prize, the 4 chosen numbers are a 4-element subset of $S$ that intersects her set in at least 2 elements.
We want the probability that the intersection is exactly 4 (i.e., the chosen set equals her set).

This is a finite combinatorial problem, not requiring calculus or advanced probability theory.

\textbf{Workflow Pattern:} \texttt{math\_single\_agent}\\
\textbf{Reason:} The problem is straightforward combinatorial probability. A single expert mathematician can solve it step-by-step without needing multiple perspectives or iterative refinement.

\textbf{Agent Design:}\\
One \texttt{MathSolver} agent with a system prompt that emphasizes:
\begin{itemize}[leftmargin=*,topsep=1pt,itemsep=0pt]
\item Understanding the problem (conditional probability, combinatorial counting).
\item Computing $P(A)$ and $P(B)$ using combinatorial reasoning.
\item Simplifying the fraction and finding $m+n$.
\item Outputting the final answer in the required boxed format.
\end{itemize}

\textbf{Workflow Flow:}\\
The single agent reads the problem, reasons through it, and outputs the final answer.

\medskip
\textless code\textgreater
\begin{lstlisting}[style=cspython]
# Math Single Agent Solver for Lottery Probability Problem
from workflow import AgentNode, Workflow, ToolRegistry

# No tools needed - pure reasoning
tool_registry = ToolRegistry()

# Create a math solver agent
math_agent = AgentNode(
    name="MathSolver",
    system_prompt=(
        "You are an expert mathematician who solves problems through careful reasoning.\n\n"

        "APPROACH:\n"
        "1. Read the problem carefully and identify what is being asked\n"
        "2. Break down the problem into smaller steps\n"
        "3. Apply relevant mathematical concepts and formulas\n"
        "4. Show your work step by step\n"
        "5. Verify your answer makes sense\n\n"

        "For this specific problem:\n"
        "- Jen picks 4 distinct numbers from S = {1,2,...,10}.\n"
        "- 4 numbers are randomly chosen from S (without replacement).\n"
        "- She wins a prize if at least 2 of her numbers match the 4 chosen numbers.\n"
        "- She wins the grand prize if all 4 of her numbers match the 4 chosen numbers.\n"
        "- We need: P(grand prize | prize) = P(grand prize) / P(prize).\n"
        "- Compute using combinatorial counting.\n\n"

        "OUTPUT FORMAT:\n"
        "- Show your complete reasoning process\n"
        "- Put your final numerical answer in \\boxed{} format\n"
        "- Example: The answer is \\boxed{42}\n\n"

        "IMPORTANT: Your final answer MUST be in \\boxed{answer} format."
    ),
    tool_registry=tool_registry,
    max_turns=1
)

# Create workflow
workflow = Workflow(name="math_single_solver")
workflow.add_node(math_agent)

# Define the question
question = (
    "Jen enters a lottery by picking 4 distinct numbers from S = {1,2,3,...,9,10}. "
    "4 numbers are randomly chosen from S. She wins a prize if at least two of her numbers "
    "were 2 of the randomly chosen numbers, and wins the grand prize if all four of her numbers "
    "were the randomly chosen numbers. The probability of her winning the grand prize given that "
    "she won a prize is m/n where m and n are relatively prime positive integers. Find m+n."
)

# Run workflow
print("================================================")
print("FINAL ANSWER:")
result = workflow.run(question)
print(result.content)
print("================================================")
\end{lstlisting}
\textless /code\textgreater
\end{designerbox}

\begin{agentbox}{MathSolver}
\textbf{Problem Type:} MATH

\textbf{Problem Analysis:}\\
This is a probability problem involving combinatorial counting. We need to find:
\begin{itemize}[leftmargin=*,topsep=1pt,itemsep=0pt]
\item $P(\text{grand prize} \mid \text{prize}) = P(\text{grand prize}) / P(\text{prize})$
\item Where:
  \begin{itemize}[leftmargin=*,topsep=0pt,itemsep=0pt]
  \item Grand prize: all 4 numbers match
  \item Prize: at least 2 numbers match
  \end{itemize}
\item We need to compute using combinatorial counting from the set $S = \{1,2,\dots,10\}$
\item Then express as $m/n$ in lowest terms and find $m+n$
\end{itemize}

\textbf{Approach:}
\begin{enumerate}[leftmargin=*,topsep=1pt,itemsep=0pt]
\item Compute total number of ways to choose 4 numbers from 10: $C(10,4)$
\item Compute $P(\text{grand prize})$: number of ways to match all 4 numbers
\item Compute $P(\text{prize})$: number of ways to match exactly 2 or exactly 3 or exactly 4 numbers
\item Use conditional probability formula: $P(\text{grand prize} \mid \text{prize}) = P(\text{grand prize}) / P(\text{prize})$
\item Simplify to lowest terms and find $m+n$
\end{enumerate}

\textbf{Solution:}

Let's compute step by step:

1. Total number of ways to choose 4 numbers from 10:\\
   \hspace*{2em}$C(10,4) = 10! / (4! \cdot 6!) = (10\times9\times8\times7)/(4\times3\times2\times1) = 210$

2. $P(\text{grand prize})$: All 4 numbers match
\begin{itemize}[leftmargin=2.5em,topsep=0pt,itemsep=0pt]
   \item There's only 1 way to choose exactly Jen's 4 numbers
   \item So $P(\text{grand prize}) = 1/210$
\end{itemize}

3. $P(\text{prize})$: At least 2 numbers match
\begin{itemize}[leftmargin=2.5em,topsep=0pt,itemsep=0pt]
   \item This includes: exactly 2 matches, exactly 3 matches, exactly 4 matches
   \item Exactly 4 matches: 1 way (already counted)
   \item Exactly 3 matches: choose 3 of Jen's numbers and 1 from the remaining 6\\
     $C(4,3) \cdot C(6,1) = 4 \cdot 6 = 24$
   \item Exactly 2 matches: choose 2 of Jen's numbers and 2 from the remaining 6\\
     $C(4,2) \cdot C(6,2) = 6 \cdot 15 = 90$
   \item So total ways for prize $= 1 + 24 + 90 = 115$
   \item $P(\text{prize}) = 115/210$
\end{itemize}

4. $P(\text{grand prize} \mid \text{prize}) = P(\text{grand prize}) / P(\text{prize}) = (1/210) / (115/210) = 1/115$

5. $1/115$ is already in lowest terms ($1$ and $115$ are coprime)
\begin{itemize}[leftmargin=2.5em,topsep=0pt,itemsep=0pt]
   \item $m = 1$, $n = 115$
   \item $m+n = 1 + 115 = 116$
\end{itemize}

\textbf{Approach:} Compute probabilities using combinatorial counting of ways to match numbers.

\textbf{Answer:} $\boxed{116}$

\textbf{Confidence:} High
\end{agentbox}

\begin{answerbox}
$\boxed{116}$\quad\textcolor{cs-elide}{\footnotesize ground truth: $116$ \cmark}
\end{answerbox}

\par\medskip\hrule height 0.4pt\par\medskip

\casestudy{2}{Disagreement resolved by an ensemble + judge}{Math (AIME 2024 \#6)}

\begin{questionbox}
Find the number of ways to place a digit in each cell of a $2\times3$ grid so
that the sum of the two numbers formed by reading left-to-right is $999$ and
the sum of the three numbers formed by reading top-to-bottom is $99$.
\[
\begin{array}{|c|c|c|}\hline 0 & 0 & 8 \\\hline 9 & 9 & 1 \\\hline\end{array}
\quad\text{(example: }8+991=999,\;9+9+81=99\text{)}
\]
\end{questionbox}

\begin{designerbox}
\textbf{Problem Type:} MATH

\textbf{Problem Analysis:}\\
We have a $2\times 3$ grid of digits (0--9).
\begin{itemize}[leftmargin=*,topsep=1pt,itemsep=0pt]
\item The two numbers formed by reading left to right (top row and bottom row) must sum to $999$.
\item The three numbers formed by reading top to bottom (first column, second column, third column) must sum to $99$.
\end{itemize}
We need to count the number of valid digit placements.

Let the grid be:
\[
\begin{array}{|c|c|c|}\hline a & b & c \\\hline d & e & f \\\hline\end{array}
\]
Then:
\begin{enumerate}[leftmargin=*,topsep=1pt,itemsep=0pt]
\item Top row number: $100a + 10b + c$
\item Bottom row number: $100d + 10e + f$
\item Condition: $(100a + 10b + c) + (100d + 10e + f) = 999$
\item First column number: $10a + d$
\item Second column number: $10b + e$
\item Third column number: $10c + f$
\item Condition: $(10a + d) + (10b + e) + (10c + f) = 99$
\end{enumerate}
We need to count integer solutions with $0 \le a,b,c,d,e,f \le 9$.

This is a combinatorial counting problem with linear constraints.

\textbf{Workflow Pattern:} \texttt{math\_ensemble\_judge}\\
\textbf{Reason:} The problem involves constraints that can be approached algebraically, combinatorially, or by case analysis. Multiple perspectives can help verify correctness, and a judge can select the most reliable answer.

\textbf{Agent Design:}
\begin{itemize}[leftmargin=*,topsep=1pt,itemsep=0pt]
\item \textbf{AlgebraicSolver:} Focus on solving the system of equations.
\item \textbf{GeometricSolver:} Think in terms of digit constraints and possible ranges.
\item \textbf{CombinatorialSolver:} Use counting arguments or case analysis.
\item \textbf{MathJudge:} Compare the three solutions and pick the best.
\end{itemize}

\textbf{Workflow Flow:}
\begin{enumerate}[leftmargin=*,topsep=1pt,itemsep=0pt]
\item Each solver independently solves the problem.
\item The judge receives all three solutions.
\item The judge evaluates correctness and selects one answer.
\item The judge outputs the final answer in $\boxed{\text{}}$ format.
\end{enumerate}

\medskip
\textless code\textgreater
\begin{lstlisting}[style=cspython]
from workflow import AgentNode, Workflow, ToolRegistry
from workflow.nodes import EnsembleNode

# No tools needed - pure reasoning
tool_registry = ToolRegistry()

# Create three math solvers with different approaches
solver1 = AgentNode(
    name="AlgebraicSolver",
    system_prompt=(
        "You are a mathematician who prefers algebraic approaches.\n"
        "Solve the problem using algebraic manipulation and equations.\n"
        "Show your reasoning step by step.\n"
        "Put your final answer in \\boxed{} format.\n\n"
        "Problem: Find the number of ways to place a digit in each cell of a 2x3 grid so that "
        "the sum of the two numbers formed by reading left to right is 999, "
        "and the sum of the three numbers formed by reading top to bottom is 99.\n"
        "Let the grid be:\n"
        "\\[\\begin{array}{|c|c|c|} \\hline a & b & c \\ \\hline d & e & f \\ \\hline \\end{array}\\]\n"
        "Then:\n"
        "1. Top row number: 100a + 10b + c\n"
        "2. Bottom row number: 100d + 10e + f\n"
        "3. Condition: (100a + 10b + c) + (100d + 10e + f) = 999\n"
        "4. First column number: 10a + d\n"
        "5. Second column number: 10b + e\n"
        "6. Third column number: 10c + f\n"
        "7. Condition: (10a + d) + (10b + e) + (10c + f) = 99\n"
        "Count integer solutions with 0 <= a,b,c,d,e,f <= 9."
    ),
    tool_registry=tool_registry,
    max_turns=1
)

solver2 = AgentNode(
    name="GeometricSolver",
    system_prompt=(
        "You are a mathematician who thinks geometrically and visually.\n"
        "Solve the problem using geometric intuition or visual reasoning.\n"
        # ... (same problem statement as solver1) ...
    ),
    tool_registry=tool_registry,
    max_turns=1
)

solver3 = AgentNode(
    name="CombinatorialSolver",
    system_prompt=(
        "You are a mathematician skilled in counting and combinatorics.\n"
        "Solve the problem using combinatorial arguments or case analysis.\n"
        # ... (same problem statement as solver1) ...
    ),
    tool_registry=tool_registry,
    max_turns=1
)

# Create judge agent to select the best answer
judge_agent = AgentNode(
    name="MathJudge",
    system_prompt=(
        "You are a senior mathematician acting as a judge.\n\n"
        "You will receive solutions from three different solvers.\n"
        "Your task is to:\n"
        "1. Review each solution carefully\n"
        "2. Check the reasoning and calculations\n"
        "3. Identify which solution is correct (or most likely correct)\n"
        "4. Select ONE answer as the final answer\n\n"
        "OUTPUT FORMAT:\n"
        "- Briefly explain which solver's answer you chose and why\n"
        "- Put the selected final answer in \\boxed{} format\n\n"
        "IMPORTANT: You must select ONE answer and output it in \\boxed{answer} format.\n\n"
        # ... (same problem statement as solver1) ...
    ),
    tool_registry=tool_registry,
    max_turns=1
)

# Create ensemble with judge selection
ensemble = EnsembleNode(
    name="MathEnsemble",
    agents=[solver1, solver2, solver3],
    strategy="consensus",
    consensus_agent=judge_agent
)

# Create workflow
workflow = Workflow(name="math_ensemble_judge")
workflow.add_node(ensemble)
\end{lstlisting}
\textless /code\textgreater

\textcolor{cs-elide}{\footnotesize\textit{(In the actual file, the three
solver \texttt{system\_prompt}s contain the same problem statement
verbatim; collapsed here as ``$\ldots$ same problem statement $\ldots$'' to
save space.)}}
\end{designerbox}

\begin{agentbox}{Solver~1 -- AlgebraicSolver \xmark}
\textbf{Problem Type:} MATH \quad
\textbf{Approach:} We have:
\begin{enumerate}[leftmargin=*,topsep=1pt,itemsep=0pt,label=(\arabic*)]
\item $100a + 10b + c + 100d + 10e + f = 999$
\item $10a + d + 10b + e + 10c + f = 99$
\end{enumerate}

We can treat these as two equations in six variables, but we can also think
of them as constraints on the digits. Let's denote: Equation~(1):
$100(a+d)+10(b+e)+(c+f)=999$. Equation~(2):
$10(a+c)+(d+e)+(b+f)=99$.

\elide[the solver continues from this rewriting of (2)]

So $X=a+d=9$, $Y=b+e=9$, $Z=c+f=9$. We also have the second equation:
$10(a+c)+(d+e)+(b+f)=99$. We can check if this is automatically satisfied
given $a+d=9$, $b+e=9$, $c+f=9$. Let's verify: from $a+d=9$, $d=9-a$. From
$b+e=9$, $e=9-b$. From $c+f=9$, $f=9-c$. Then $a+c$ is just $a+c$.
$d+e=(9-a)+(9-b)=18-a-b$. $b+f=b+(9-c)=9+b-c$. So
\[
10(a+c)+(18-a-b)+(9+b-c)\;=\;9a + 9c + 27.
\]
We want this to equal $99$: $9a+9c=72$, so $a+c=8$.

Possible $(a,c)$: $(0,8),(1,7),\dots,(8,0)$ -- $9$ possibilities.
For each $b$ from $0$ to $9$, $e=9-b$ -- $10$ possibilities.
Thus total $= 9 \times 10 = 90$.

\textbf{Answer:} $\boxed{90}$ \quad \textbf{Confidence:} High
\end{agentbox}

\noindent\textcolor{cs-elide}{\footnotesize\textit{Annotation: this solver
miscopied equation~(2) as $10(a{+}c) + (d{+}e) + (b{+}f) = 99$ instead of
the actual $10(a{+}b{+}c) + (d{+}e{+}f) = 99$ -- the $b$ coefficient is
lost. This propagates to the wrong count of $90$.}}

\medskip

\begin{agentbox}{Solver~2 -- GeometricSolver \cmark}
\textbf{Problem Type:} MATH

\textbf{Geometric/Visual Reasoning Approach:}\\
Let $A=a+d$, $B=b+e$, $C=c+f$. Then~(1) becomes $100A+10B+C=999$. Since
$A,B,C\in[0,18]$, we get $A=9, B=9, C=9$, hence
\[
a+d=9,\quad b+e=9,\quad c+f=9.
\]
Now from~(2): $10(a+b+c)+(d+e+f)=99$. But
$d+e+f=(9-a)+(9-b)+(9-c)=27-(a+b+c)$. Substitute:
\[
10(a+b+c)+27-(a+b+c)=99
\;\Longrightarrow\;9(a+b+c)=72
\;\Longrightarrow\;a+b+c=8.
\]
That's a stars-and-bars count of nonnegative integer solutions to
$a+b+c=8$ with $0\le a,b,c\le 9$. Since $8\le 9$, the upper bound is not
restrictive, so the number is $\binom{8+3-1}{3-1}=\binom{10}{2}=45$.

\textbf{Answer:} $\boxed{45}$ \quad \textbf{Confidence:} High
\end{agentbox}

\begin{agentbox}{Solver~3 -- CombinatorialSolver (no clean answer)}
\textbf{Problem Type:} MATH

\textbf{Problem Analysis:}\\
We have a $2\times 3$ grid with digits $a, b, c$ in the top row and
$d, e, f$ in the bottom row, each from $0$ to $9$. Conditions:
\begin{enumerate}[leftmargin=*,topsep=1pt,itemsep=0pt,label=(\arabic*)]
\item $(100a + 10b + c) + (100d + 10e + f) = 999$
\item $(10a + d) + (10b + e) + (10c + f) = 99$
\end{enumerate}
We need to count integer solutions $(a,b,c,d,e,f)$ with $0\le a,b,c,d,e,f\le 9$.

This is a \textbf{combinatorial counting problem} with constraints. We can
approach by interpreting the constraints as digit-wise equations (mod 10
and carry-over), using case analysis on possible carries, or solving the
system algebraically and counting integer solutions.

\textbf{Approach:} we will treat the two conditions as digit-wise
constraints with possible carries. [\,\dots\,]

\elide[long case analysis on $c{+}f\in\{9,19\}$, $b{+}e\in\{9,19\}$,
$a{+}d\in\{9,19\}$ derives mutually inconsistent constraints]

So $S_2 = 9$, $S_3 = 9$. But $S_2 = b+e = 9$, $S_3 = c+f = 9$. Then
$a+b+c = 9$ and $a+b+c = 18$. So we must have a carry from the tens
place? Let's think: we have $10S_2 + S_3 = 99$. So $S_2 = 9$, $S_3 = 9$.
But $S_2 = b+e = 9$, $S_3 = c+f = 9$. Then $a+b+c = 9$ and $a+b+c = 18$.
So we must have a carry from the tens place? Let's think: we have
$10S_2 + S_3 = 99$. So $S_2 = 9$, $S_3 = 9$. But $S_2 = b+e = 9$,
$S_3 = c+f = 9$. Then $a+b+c = 9$ and $a+b+c =$
\elide[the model loops on this passage until the token budget is exhausted; no
\textbackslash boxed\{\} answer is emitted]
\end{agentbox}

\begin{judgebox}
\textbf{Key Insight:}\\
We can derive constraints from the equations, define $X=a+d, Y=b+e, Z=c+f$,
solve $100X+10Y+Z=999$ to get $X=9, Y=9, Z=9$, then use the column equation
$10(a+b+c)+(d+e+f)=99$ together with $d+e+f=27-(a+b+c)$ to get
$a+b+c=8$. Counting nonnegative integer solutions to $a+b+c=8$ with the
unrestrictive bound $a,b,c\le 9$:
$\binom{8+3-1}{3-1} = \binom{10}{2} = 45$.

\textbf{Reason for selection:} Solution~2 (GeometricSolver) correctly
derived the constraints and counted the solutions. Solution~1
(AlgebraicSolver) arrives at $90$ but that is incorrect because it
misinterprets the second condition. Solution~3 (CombinatorialSolver) has
a flawed reasoning chain with repeated contradictions.

\textbf{Answer:} $\boxed{45}$ \quad \textbf{Confidence:} High
\end{judgebox}

\begin{answerbox}
$\boxed{45}$\quad\textcolor{cs-elide}{\footnotesize ground truth: $45$
\cmark. The judge correctly overrides Solver~1's mistake and the
unfinished Solver~3.}
\end{answerbox}

\par\medskip\hrule height 0.4pt\par\medskip

\casestudy{3}{Array eversion via reflection (solver $+$ critic)}{Code (CodeContests \#46)}

\begin{questionbox}
\textbf{Array eversion.} You are given an array $a$ of length $n$. An
\emph{eversion} takes the last element $x = a_n$, partitions $a$ into a
left part (elements $\le x$, in order) and a right part (elements $> x$,
in order), and replaces $a$ with their concatenation. Output the minimum
number of eversions $k$ such that the array stops changing after $k$
eversions.\\
\emph{Example:} $[2,4,1,5,3]\to[2,1,3,4,5]$, then unchanged: $k=1$.\\
Constraints: $1\le n\le 2{\cdot}10^5$, $1\le a_i\le 10^9$.
\end{questionbox}

\begin{designerbox}
\textbf{Problem Type:} CODE

\textbf{Problem Analysis:}\\
This problem asks us to find the minimum number of "eversions" needed for
an array to stop changing. An eversion operation:
\begin{enumerate}[leftmargin=*,topsep=1pt,itemsep=0pt]
\item Takes the last element \texttt{x = a\_n}
\item Partitions the array into two stable parts: elements $\le x$ (left)
      and elements $> x$ (right)
\item Concatenates left $+$ right
\end{enumerate}
The process stops when an eversion doesn't change the array. We need to
compute the number of eversions until this happens.

\textbf{Key observations:}
\begin{itemize}[leftmargin=*,topsep=1pt,itemsep=0pt]
\item Each eversion moves the last element to the end of the left
      partition (all elements $\le$ it)
\item The array eventually becomes sorted in non-decreasing order
\item The process stops when the array is already sorted (since then the
      last element is the maximum, and all elements $\le$ it)
\item We need to count how many eversions it takes to reach a sorted array
\end{itemize}

\textbf{Approach:}\\
We can simulate the process, but for $n$ up to $2\times10^5$, we need an
efficient algorithm. Let's think about what happens:
\begin{itemize}[leftmargin=*,topsep=1pt,itemsep=0pt]
\item The last element in the current array is moved to the end of the
      left partition
\item This is equivalent to: find the last element that is $\le$ the
      current last element, and move it to the end
\item Actually, the eversion operation is: take the last element \texttt{x},
      move all elements $\le x$ to the front (preserving order), then
      append the remaining elements
\end{itemize}
We can think of it as: after each eversion, the array becomes
\texttt{[all elements $\le x$] + [all elements $> x$]} where \texttt{x} is
the original last element. We need to find how many such operations until
the array is sorted.

\textbf{Workflow Pattern:} \texttt{code\_solver\_unittest}\\
This problem has subtle behaviour (stable partition, counting eversions).
A single solver might miss edge cases. The reflection pattern with a
\texttt{UnitTestAgent} will ensure thorough testing.

\textbf{Agent Design:}
\begin{enumerate}[leftmargin=*,topsep=1pt,itemsep=0pt]
\item \textbf{CodeSolver}: implements the solution, tests with examples
\item \textbf{UnitTestAgent}: writes comprehensive tests including edge cases
\end{enumerate}

\textbf{Workflow Flow:}
\begin{enumerate}[leftmargin=*,topsep=1pt,itemsep=0pt]
\item CodeSolver writes initial solution
\item UnitTestAgent tests it thoroughly
\item CodeSolver refines based on feedback
\item UnitTestAgent verifies again
\item Final solution output
\end{enumerate}

\medskip
\textless code\textgreater
\begin{lstlisting}[style=cspython]
from workflow import AgentNode, Workflow, ToolRegistry
from workflow.nodes import ReflectionNode
import subprocess

# Set up code interpreter tool
tool_registry = ToolRegistry()

def execute_code(code: str) -> str:
    """Execute Python code and return stdout/stderr."""
    try:
        result = subprocess.run(
            ["python", "-c", code],
            capture_output=True, text=True, timeout=30
        )
        output = ""
        if result.stdout:
            output += f"stdout:\n{result.stdout}\n"
        if result.stderr:
            output += f"stderr:\n{result.stderr}\n"
        return output or "(no output)"
    except subprocess.TimeoutExpired:
        return "Error: Code execution timed out (30s limit)"

tool_registry.register(
    name="execute_code",
    func=execute_code,
    description="Execute Python code and return stdout/stderr",
    parameters={
        "type": "object",
        "properties": {
            "code": {"type": "string", "description": "Python code to execute"}
        },
        "required": ["code"]
    }
)

# Solver agent writes the solution
solver_agent = AgentNode(
    name="CodeSolver",
    system_prompt=(
        "You are an expert programmer. Solve the coding problem step by step.\n\n"
        "PROBLEM: Given an array a of length n, perform eversions until the array stops changing. "
        "An eversion: take last element x, partition array into elements <= x (left) and > x (right), "
        "concatenate left + right. Find minimum k such that after k eversions the array stops changing.\n\n"
        "You have access to an `execute_code` tool. Use it to test your solution:\n"
        '<tool_call>{"name": "execute_code", "arguments": {"code": "your_python_code_here"}}</tool_call>\n\n'
        "WORKFLOW:\n"
        "1. Understand the problem and constraints\n"
        "2. Write your solution\n"
        "3. Test it with the provided examples using execute_code\n"
        "4. If you receive feedback from a tester, fix the identified issues\n"
        "5. Wrap your final solution in <solution>...</solution> tags\n"
    ),
    tool_registry=tool_registry,
    max_turns=5
)

# Unit test agent writes and runs comprehensive tests
critic_agent = AgentNode(
    name="UnitTestAgent",
    system_prompt=(
        "You are an expert software tester. Your job is to verify a proposed solution "
        "by writing and running comprehensive unit tests.\n\n"
        "ORIGINAL PROBLEM: Given an array a of length n, perform eversions until the array stops changing. "
        "...\n\n"
        "WORKFLOW:\n"
        "1. Read the ORIGINAL PROBLEM and the proposed solution\n"
        "2. Write comprehensive test cases covering all examples + edge cases\n"
        "3. Run the tests using execute_code\n"
        "4. Report results clearly\n\n"
        "OUTPUT FORMAT:\n"
        "- List each test case and its result (PASS/FAIL)\n"
        "- If any test fails, explain what went wrong\n"
        "- Final verdict: CORRECT or INCORRECT\n"
    ),
    tool_registry=tool_registry,
    max_turns=5
)

# Solver -> Tester -> Solver refines -> Tester -> final
reflection = ReflectionNode(
    name="CodeSolverTestReflection",
    agent=solver_agent,
    critic_agent=critic_agent,
    num_iterations=2
)

workflow = Workflow(name="code_solver_unittest")
workflow.add_node(reflection)
\end{lstlisting}
\textless /code\textgreater
\end{designerbox}

\begin{agentbox}{CodeSolver -- iteration 1}
I need to understand the problem better. Let me think about the eversion
operation:
\begin{enumerate}[leftmargin=*,topsep=1pt,itemsep=0pt]
\item Take the last element \texttt{x}
\item Partition array into elements $\le x$ (left) and $> x$ (right),
      maintaining original order
\item Concatenate left $+$ right
\end{enumerate}
We need to find the minimum $k$ such that after $k$ eversions the array
stops changing.

Let me test with the example. First example: $[2,4,1,5,3]$. Last element
is 3. Partition: elements $\le 3$ are $[2,1,3]$, elements $> 3$ are
$[4,5]$. Result: $[2,1,3,4,5]$.

\elide[similar walk-through of the second and third examples]

Wait, let me think: when we partition, the last element $x$ is in the
left partition (since it's $\le x$). So after partitioning, $x$ will be at
the end of the left partition. Then we concatenate left + right. So the
last element of the array after eversion is the last element of the left
partition, which is $x$. So the last element remains the same!

That means the array will stop changing when the last element is the
maximum element in the array. \emph{Let me write a function to simulate
the eversion and see what's happening.}

\medskip
\textbf{Tool call:} \texttt{execute\_code} with a brute-force simulator.
\end{agentbox}

\begin{toolbox}
stdout:\\
First example: \texttt{Original: [2, 4, 1, 5, 3]}; \texttt{k: 1}\\
Second example: \texttt{Original: [5, 3, 2, 4, 1]}; \texttt{k: 2}\\
Third example: \texttt{Original: [1, 1, 1, 1]}; \texttt{k: 0}
\end{toolbox}

\begin{judgebox}
\textbf{(UnitTestAgent)}\\
I'll analyze the response to the problem. Let me first understand the
problem and then critique the proposed solution.

\textbf{Understanding the Problem:}\\
We need to find the minimum number of eversions (partition operation)
until the array stops changing. An eversion takes the last element $x$,
partitions the array into elements $\le x$ (left) and $>x$ (right), then
concatenates left $+$ right.

\textbf{Key Insight from the Response:}\\
The response correctly identifies that the last element changes in each
eversion. Specifically, after an eversion, the new last element becomes
the last element of the right partition (elements $>x$). This means we
need to find how many steps it takes for the last element to become the
maximum element in the array.

\textbf{Critique of the Response:}
\begin{enumerate}[leftmargin=*,topsep=1pt,itemsep=0pt]
\item \textbf{Logical Flaw:} the response has a critical error in its
      reasoning. It claims that ``if current $<M$, then $M$ is the
      maximum element $>$ current'', but this is not true. In the second
      example, when current $= 1$, $M = 5$, but the maximum element
      $>1$ is $4$, not $5$. So the next current should be $4$, not $M$.
\item \textbf{Incomplete Algorithm:} the response doesn't provide a
      complete algorithm. It only describes the problem and the insight
      but doesn't give a working solution.
\item \textbf{Missing Edge Cases:} the response doesn't consider edge
      cases like duplicate elements, already sorted arrays, or arrays
      where all elements are equal.
\item \textbf{No Code Implementation:} the response doesn't provide any
      code implementation, making it impossible to test or use.
\item \textbf{Confusion in Terminology:} the response uses ``max of
      elements $>$ current'' but doesn't explain how to compute this
      efficiently.
\end{enumerate}

\textbf{Let me write a complete solution and test it:}

\elide[\texttt{execute\_code} tool call with a candidate \texttt{min\_eversions}
function and tests on the three sample inputs]
\end{judgebox}

\begin{agentbox}{CodeSolver -- final solution}
I'll provide the final solution with a clear explanation.

\textbf{Approach:} The problem requires finding the minimum number of
eversions until the array stops changing. The key insight is that after
each eversion, the new last element is the last element of the right
partition (elements $> x$). This means we need to find how many steps it
takes for the last element to become the maximum element in the array.

\textbf{Algorithm:}
\begin{enumerate}[leftmargin=*,topsep=1pt,itemsep=0pt]
\item Find the maximum element $M$ in the array.
\item If the last element is already $M$, return $0$.
\item Otherwise, find the number of steps where each step moves to the
      last element in the array that is greater than the current element.
\item This can be done efficiently by precomputing the maximum elements
      in decreasing order from the end of the array.
\end{enumerate}

\textbf{Solution:}

\begin{lstlisting}[style=cspython]
def solve():
    import sys

    input = sys.stdin.read
    data = input().split()

    t = int(data[0])
    idx = 1
    results = []

    for _ in range(t):
        n = int(data[idx]); idx += 1
        a = list(map(int, data[idx:idx + n])); idx += n

        # Find the maximum element
        M = max(a)

        # If the last element is already the maximum, no eversions needed
        if a[-1] == M:
            results.append(0)
            continue

        # Precompute the suffix-maxima in decreasing order
        max_elements = []
        for i in range(n - 1, -1, -1):
            if not max_elements or a[i] > max_elements[-1]:
                max_elements.append(a[i])

        # Walk the chain until current == M
        current = a[-1]
        k = 0
        while current != M:
            next_val = -1
            for val in max_elements:
                if val > current:
                    next_val = val
                    break
            if next_val == -1:
                break
            current = next_val
            k += 1

        results.append(k)

    sys.stdout.write("\n".join(map(str, results)))

if __name__ == "__main__":
    solve()
\end{lstlisting}

\textbf{Confidence:} High.
\end{agentbox}

\begin{answerbox}
\textbf{Tests passed: 1/1.} Submitted solution accepted (\cmark). The
critic caught the missing implementation in iteration 1 and the solver
returned a clean final program in iteration 2.
\end{answerbox}


\subsection{A Case Study Where Reflection Repairs the Solver}
\label{sec:appendix-sft-to-rl-reflection-case}

{\small
We compare the same held-out AIME 2024 example under the SFT cold-start
model and the RL checkpoint. This case is more diagnostic
than a simple final-answer comparison: the RL trajectory first reaches an
impossible geometric constraint, the critic identifies the faulty distance
model, and the refined solver replaces it with a valid similar-triangles
equation. In contrast, the SFT workflow uses an \texttt{ensemble\_judge}
structure, but the judge repeats the same invalid angle-packing assumption
and no valid final answer is extracted.
\par}

\begin{questionbox}
Eight circles of radius $34$ are sequentially tangent, and two of the circles
are tangent to $AB$ and $BC$ of triangle $ABC$, respectively. $2024$ circles
of radius $1$ can be arranged in the same manner. The inradius of triangle
$ABC$ can be expressed as $\frac{m}{n}$, where $m$ and $n$ are relatively
prime positive integers. Find $m+n$.
\end{questionbox}

\begin{answerbox}
\textbf{Ground truth:} $197$.
\end{answerbox}

\casestudy{SFT}{Ensemble + judge adds breadth but does not repair the model}{Math / circle packing}

\begin{designerbox}
\textbf{Model:} SFT model\\
\textbf{Generated workflow pattern:} \texttt{ensemble\_judge}\\
\textbf{Executor agents:} \texttt{AlgebraicSolver}, \texttt{GeometricSolver},
\texttt{CombinatorialSolver}, \texttt{MathJudge}\\
\textbf{Evaluation result:} \xmark\; no valid final answer extracted.

\medskip
\textbf{Generated workflow excerpt (verbatim):}
\begin{lstlisting}[style=cspython]
from workflow import AgentNode, Workflow, ToolRegistry
from workflow.nodes import EnsembleNode

tool_registry = ToolRegistry()

solver1 = AgentNode(
    name="AlgebraicSolver",
    system_prompt=(
        "You are a mathematician who prefers algebraic approaches.\n"
        "Solve the problem using algebraic manipulation and equations.\n"
        "Consider the geometry: circles tangent to two sides of an angle and to each other.\n"
        "Derive relationships between the triangle's inradius, the circle radii, and the number of circles.\n"
        "Show your reasoning step by step.\n"
        "Put your final answer in \\boxed{} format."
    ),
    tool_registry=tool_registry,
    max_turns=1
)

solver2 = AgentNode(
    name="GeometricSolver",
    system_prompt=(
        "You are a mathematician who thinks geometrically and visually.\n"
        "Solve the problem using geometric intuition.\n"
        "Imagine the circles packed in the angle at vertex B of triangle ABC.\n"
        "Use properties of tangents, similar triangles, and circle arrangements.\n"
        "Show your reasoning step by step.\n"
        "Put your final answer in \\boxed{} format."
    ),
    tool_registry=tool_registry,
    max_turns=1
)

solver3 = AgentNode(
    name="CombinatorialSolver",
    system_prompt=(
        "You are a mathematician skilled in counting and patterns.\n"
        "Solve the problem by analyzing the pattern of circle arrangement.\n"
        "Note that the same pattern works for radii 34 and 1, suggesting a scaling factor.\n"
        "Relate the number of circles to the inradius through the geometry of the packing.\n"
        "Show your reasoning step by step.\n"
        "Put your final answer in \\boxed{} format."
    ),
    tool_registry=tool_registry,
    max_turns=1
)

judge_agent = AgentNode(
    name="MathJudge",
    system_prompt=(
        "You are a senior mathematician acting as a judge.\n\n"
        "You will receive solutions from three different solvers.\n"
        "Your task is to:\n"
        "1. Review each solution carefully\n"
        "2. Check the reasoning and calculations\n"
        "3. Identify which solution is correct (or most likely correct)\n"
        "4. Select ONE answer as the final answer\n\n"
        "IMPORTANT: You must select ONE answer and output it in \\boxed{answer} format."
    ),
    tool_registry=tool_registry,
    max_turns=1
)

ensemble = EnsembleNode(
    name="MathEnsemble",
    agents=[solver1, solver2, solver3],
    strategy="consensus",
    consensus_agent=judge_agent
)

workflow = Workflow(name="math_ensemble_judge")
workflow.add_node(ensemble)
\end{lstlisting}
\end{designerbox}

\begin{agentbox}{SFT GeometricSolver output}
\begin{lstlisting}[style=cstool]
**Problem Type:** MATH

**Problem Analysis:** This is a geometry problem involving circle packing in an angle. We have triangle ABC with vertex B. At vertex B, we pack circles in the angle formed by sides AB and BC.

The key insight is that the packing pattern in an angle is determined by the angle itself. The number of circles that can fit in an angle depends on the angle's measure. Specifically, for circles of radius r packed in an angle theta, the number N of circles that can fit is approximately
N = floor(pi/(2r) * tan(theta/2))
or something similar, but here we have exact counts given, so we can set up an equation.

For R = 34, N = 8
For R = 1, N = 2024

Thus:
pi/(2*34) * tan(theta/2) = 8
pi/(2*1) * tan(theta/2) = 2024

Divide the second equation by the first:
(pi/(2*1) * tan(theta/2)) / (pi/(2*34) * tan(theta/2)) = 2024/8
34 = 253

So 34 = 253? That's not true. Wait, that suggests my formula is wrong.

\elide[the solver continues trying variants of the same angle-packing model]

This is too complex. Let me try a different approach: the packing in an angle is similar to circles in a wedge. The number of circles that can fit in an angle theta with radius r is given by
N = floor(pi/(2r) * tan(theta/2)).
But we have exact counts, so we can write:
For r = 34, N = 8
For r = 1, N = 2024

Thus:
pi/(2*34) * tan(theta/2) = 8
pi/(2*1) * tan(theta/2) = 2024

Divide:
34 = 253, impossible. So my formula is wrong.
\end{lstlisting}
\end{agentbox}

\begin{judgebox}
\textbf{SFT judge output (verbatim excerpt).}
\begin{lstlisting}[style=cstool]
**Problem Type:** MATH

**Problem Analysis:** This is a geometry problem involving circle packing in an angle. We have:
- Triangle ABC with inradius r (to be found)
- Two configurations:
  - 8 circles of radius 34 arranged sequentially tangent to each other and tangent to sides AB and BC respectively
  - 2024 circles of radius 1 arranged in the same pattern

Key insight: The arrangement likely consists of circles packed along the angle bisector of angle ABC, each tangent to both sides AB and BC, with adjacent circles tangent to each other.

\elide[the judge copies the incorrect branch rather than rejecting it]

Now apply to both cases:
1. R = 34, N = 8:
   r = 34 * (8 - sin(theta/2)) * cos(theta/2)
2. R = 1, N = 2024:
   r = 1 * (2024 - sin(theta/2)) * cos(theta/2)

Since r is the same, we can set them equal:
34 * (8 - sin(theta/2)) * cos(theta/2) =
(2024 - sin(theta/2)) * cos(theta/2)

Assuming cos(theta/2) != 0, we can divide:
34 * (8 - sin(theta/2)) = 2024 - sin(theta/2)
272 - 34 sin(theta/2) = 2024 - sin(theta/2)
272 - 2024 = 34 sin(theta/2) - sin(theta/2)
-1752 = 33 sin(theta/2)
sin(theta/2) = -1752 / 33 = -53.0909...

This is impossible because sin(theta/2) must be positive (0 < theta/2 < 90 degrees). So my model is still wrong.
\end{lstlisting}

\textbf{Failure mode.}
The SFT workflow samples multiple solver roles, but none of them provides a
mechanism for forcing revision after the contradiction is found. The judge
repeats the invalid model instead of rejecting it, so the transcript ends
without a boxed valid answer.
\end{judgebox}

\casestudy{RL}{Reflection turns the contradiction into a corrected derivation}{Math / circle packing}

\begin{designerbox}
\textbf{Model:} RL checkpoint \\
\textbf{Generated workflow pattern:} \texttt{reflection}\\
\textbf{Executor agents:} \texttt{MathSolver} + \texttt{MathCritic}\\
\textbf{Evaluation result:} \cmark\; extracted answer $197$.

\medskip
\textbf{Generated workflow excerpt (verbatim):}
\begin{lstlisting}[style=cspython]
from workflow import AgentNode, Workflow, ToolRegistry
from workflow.nodes import ReflectionNode

tool_registry = ToolRegistry()

solver_agent = AgentNode(
    name="MathSolver",
    system_prompt=(
        "You are an expert mathematician specializing in geometry and circle packing.\n\n"
        "APPROACH:\n"
        "1. Understand the configuration: triangle ABC with inradius r.\n"
        "2. There are two sequences of circles: 8 circles of radius 34, and 2024 circles of radius 1.\n"
        "3. Each sequence is tangent to two sides of the triangle and to each other.\n"
        "4. The centers of the circles in each sequence lie on a line parallel to the third side of the triangle.\n"
        "5. The distance between centers of consecutive circles is 2*radius.\n"
        "6. The inradius r is related to the distance from the incenter to the side, and the distance from the incenter to the line of centers.\n"
        "7. Set up equations using the given numbers 8, 34, 2024, 1.\n"
        "8. Solve for r, express as m/n, then compute m+n.\n\n"
        "If you receive feedback from a critic, carefully consider their points and:\n"
        "- Fix any errors in geometry or algebra.\n"
        "- Clarify any ambiguous steps.\n"
        "- Provide a corrected solution.\n\n"
        "OUTPUT: Show your complete reasoning step by step, then put your final answer in \\boxed{} format."
    ),
    tool_registry=tool_registry,
    max_turns=1
)

critic_agent = AgentNode(
    name="MathCritic",
    system_prompt=(
        "You are a meticulous math reviewer who checks geometry solutions for errors.\n\n"
        "YOUR TASK:\n"
        "1. Verify the geometric configuration: circles tangent to two sides of a triangle and to each other.\n"
        "2. Check the assumption that centers are collinear and equally spaced.\n"
        "3. Verify the relationship between the inradius, the distance from the incenter to the line of centers, and the number of circles.\n"
        "4. Check the algebraic manipulation when solving for r.\n"
        "5. Ensure the final expression for r is in lowest terms m/n, and m+n is computed correctly.\n\n"
        "Be thorough but constructive in your criticism."
    ),
    tool_registry=tool_registry,
    max_turns=1
)

reflection = ReflectionNode(
    name="SolverCriticReflection",
    agent=solver_agent,
    critic_agent=critic_agent,
    num_iterations=2
)

workflow = Workflow(name="math_solver_critic")
workflow.add_node(reflection)
\end{lstlisting}
\end{designerbox}

\begin{agentbox}{RL MathSolver first attempt}
\begin{lstlisting}[style=cstool]
**Problem Type:** MATH

**Problem Analysis:** This is a geometry problem involving circle packing in a triangle. We have:
- Triangle ABC with inradius r
- Two sequences of circles:
  1. 8 circles of radius 34, each tangent to two sides of the triangle and to each other
  2. 2024 circles of radius 1, arranged similarly
- Each sequence's centers lie on a line parallel to the third side
- The distance between centers of consecutive circles in a sequence is 2*radius

**Approach:**
1. Let the triangle have sides a, b, c with inradius r
2. For the sequence of circles tangent to sides AB and BC, the line of centers is parallel to side AC
3. The distance from the incenter to side AC is r
4. The distance from the incenter to the line of centers of the 34-radius circles is r - 34 (or r - 1 for the 1-radius circles)
5. The distance between centers of consecutive circles in a sequence is 2*radius

\elide[the first attempt derives equations from the wrong distance relation]

But we also know that this length is (n-1) * 2r1. So:

L * (r - d)/r = (n-1) * 2r1

But d = r - r1, so:

L * (r - (r - r1))/r = (n-1) * 2r1
L * r1/r = (n-1) * 2r1
L/r = 2(n-1)
L = 2r(n-1)

So L = 2r(n-1). But this must be the same for both sequences, so:

2r(8-1) = 2r(2024-1)
14r = 4046r
14 = 4046

This is impossible, so my assumption that the line of centers is at distance r - r1 from AC is wrong.
\end{lstlisting}
\end{agentbox}

\begin{judgebox}
\textbf{RL MathCritic feedback (verbatim excerpt).}
\begin{lstlisting}[style=cstool]
**Critique of the Response:**

**1. Errors and Inaccuracies:**
- **Misunderstanding of Geometry Configuration:** The problem states "eight circles of radius 34 are sequentially tangent, and two of the circles are tangent to AB and BC of triangle ABC, respectively." This suggests there are two sequences of circles: one tangent to AB and BC (and the other two sides?), and another sequence of 2024 circles of radius 1 arranged similarly.
- **Incorrect Distance Assumption:** The response incorrectly assumes the line of centers is at distance r - r1 from side AC, but this leads to a contradiction. The correct distance from the line of centers to side AC should be derived from the geometry of the triangle and the circle tangency conditions.
- **Algebraic Error:** The derivation leads to an impossible equation (14 = 4046), indicating a fundamental flaw in the geometric model.

**4. Specific Improvements:**
- **Clarify the Geometry:** There are two sequences of circles: one tangent to AB and BC, and another tangent to AB and AC (or BC and AC). Each sequence has its own line of centers parallel to the third side.
- **Correct Distance Calculation:** The distance from the line of centers to the third side should be derived using similar triangles or coordinate geometry.
- **Set Up Correct Equations:** For each sequence, relate the number of circles, their radii, and the inradius using the geometry of the triangle.
- **Solve for Inradius:** Use the two sequences to set up two equations and solve for the inradius.
\end{lstlisting}
\end{judgebox}

\begin{agentbox}{RL MathSolver refined output after reflection}
\begin{lstlisting}[style=cstool]
**Improved Solution:**

Let's denote:
- For the 34-radius circles: n1 = 8, r1 = 34
- For the 1-radius circles: n2 = 2024, r2 = 1

Consider the sequence of circles tangent to sides AB and BC. Their centers lie on a line parallel to side AC. Let L be the length of side AC.

The length of the segment parallel to AC at distance d from AC is
L * (r - d)/r.
This length should be equal to (n-1) * 2r1 because the distance between consecutive centers is 2r1 and there are (n-1) gaps between n circles.

So we have:
L * (r - r1)/r = (n-1) * 2r1

For the first sequence (r1 = 34, n = 8):
L * (r - 34)/r = 7 * 2 * 34
L * (r - 34)/r = 476

For the second sequence (r1 = 1, n = 2024):
L * (r - 1)/r = 2023 * 2 * 1
L * (r - 1)/r = 4046

Now we have two equations:
1) L * (r - 34)/r = 476
2) L * (r - 1)/r = 4046

Divide equation 2 by equation 1:
(L * (r - 1)/r) / (L * (r - 34)/r) = 4046/476
(r - 1)/(r - 34) = 4046/476

Simplify 4046/476:
Both divisible by 2: 2023/238.
238 = 2 * 7 * 17; 2023 = 7 * 17^2.
So 2023/238 = 17/2.

So:
(r - 1)/(r - 34) = 17/2
2(r - 1) = 17(r - 34)
2r - 2 = 17r - 578
576 = 15r
r = 576/15 = 192/5

So the inradius r = 192/5. This is in simplest form (192 and 5 are coprime). So m = 192, n = 5, m+n = 197.

**Answer:** \boxed{197}

**Confidence:** High
\end{lstlisting}
\end{agentbox}

\begin{judgebox}
\textbf{Migration signal.}
The useful behavior is not just that RL selects a different team size. The
reflection workflow routes the contradiction back through a critic, and the
solver's final pass changes the mathematical model. The repair is visible in
the exact equations: the failed first attempt collapses to $14=4046$, while
the refined attempt uses $L(r-\rho)/r=2\rho(n-1)$, giving
$r=192/5$ and $m+n=197$.
\end{judgebox}

\newpage
\end{document}